\documentclass[journal]{IEEEtran}

\usepackage[american]{babel}
\usepackage{cite}
\usepackage{mathtools}
\usepackage{booktabs}
\usepackage{tikz}

\usepackage{graphicx, times}
\usepackage{subcaption} 
\usepackage{bm}
\usepackage{url}
\usepackage{amsthm, amssymb, amsfonts, amsmath}
\usepackage{bbold}
\usepackage{dsfont}
\usepackage{ragged2e}
\usepackage{booktabs}
\usepackage{algorithm, algorithmic}
\usepackage{pgfplots}
\usepackage{comment}
\usepackage{multirow}
\pgfplotsset{compat=1.17}
\usepackage{thmtools}

\newtheorem{theorem}{Theorem}
\newcommand{\repeattheorem}[2]{\noindent\textbf{Theorem~\ref{#1}.} \textit{#2}}

\newcommand{\bd}[1]{\boldsymbol{#1}}
\newcommand{\set}[1]{\mathcal{#1}}
\newcommand{\X}{\mathcal{X}}
\newcommand{\Y}{\mathcal{Y}}
\newcommand{\1}[1]{\mathds{1}(#1)}

\newcommand{\x}{\boldsymbol{x}}
\newcommand{\tht}{\boldsymbol{\theta}}
\newcommand{\Real}{\mathds{R}}

\begin{document}
%
\title{Decentralized Federated Learning of Probabilistic Generative Classifiers}

\author{Aritz Pérez, Carlos Echegoyen and Guzmán Santafé
\thanks{Aritz Pérez is at the Basque Center for Applied Mathematics, 48009 Bilbao, Spain. Email: aperez@bcamath.org}
\thanks{Carlos Echegoyen and Guzmán Santafé are with the Spatial Statistics Group and INAMAT$^2$, Public University of Navarre, 31006 Pamplona, Spain. Email:\{carlos.echegoyen, guzman.santafe\}@unavarra.es}
}

\maketitle

\begin{abstract}
Federated learning is a paradigm of increasing relevance in real world applications, aimed at building a global model across a network of heterogeneous users without requiring the sharing of private data. We focus on model learning over decentralized architectures, where users collaborate directly to update the global model without relying on a central server. In this context, the current paper proposes a novel approach to collaboratively learn probabilistic generative classifiers with a parametric form. The framework is composed by a communication network over a set of local nodes, each of one having its own local data, and a local updating rule. The proposal involves sharing local statistics with neighboring nodes, where each node aggregates the neighbors' information and iteratively learns its own local classifier, which progressively converges to a global model. Extensive experiments demonstrate that the algorithm consistently converges to a globally competitive model across a wide range of network topologies, network sizes, local dataset sizes, and extreme non-i.i.d. data distributions. 

\begin{IEEEkeywords}
Decentralized federated learning, supervised classification, empirical risk minimization, probabilistic models, non-i.i.d. data.
\end{IEEEkeywords}

\end{abstract}

\section{Introduction}
In recent years, federated learning (FL) \cite{mcmahan17, Yang19} has gained increasing attention from both the research community \cite{Jakub16, Sun23} and private companies \cite{mcmahan2020communication, Granqvist24}, as it enables the development of machine learning models across multiple users without requiring data centralization. This design inherently offers a fundamental layer of privacy while reducing the costs associated with massive data storage. FL traditionally achieves this by using a user-server architecture, where users train local models and share updates with a central server that aggregates them to build a global model \cite{Li20, Kairouz21}. In contrast, decentralized FL~\cite{lalitha2019decentralized, kalra2023, Sun23} eliminates the need for a central server by enabling users to communicate directly and collaboratively train machine learning models. This decentralized paradigm not only removes key risks associated with centralization, such as single points of failure and server communication bottlenecks, but also enhances efficiency and robustness through direct peer-to-peer communication. Moreover, it allows the system to operate over existing network topologies, reducing the need for centralized infrastructure.

The network architecture has a crucial impact on the model convergence \cite{Sun23}. In centralized FL, the central server coordinates the aggregation of local models, ensuring a direct and consistent flow of information between users and the server. However, in decentralized architectures, nodes rely on peer-to-peer communication, which entails additional challenges. The network topology plays a critical role in determining how quickly and effectively local models converge to a global solution. For example, highly connected networks facilitate faster convergence by enabling rapid information exchange among neighbors, whereas sparsely connected networks may lead to delayed updates and slower convergence.

Beyond the network topology, FL faces challenges across multiple dimensions, including handling heterogeneous data distributions, optimizing communication, managing security risks, resource-constrained environments, or ensuring fair learning, among others \cite{BANABILAH2022}. A fundamental methodological challenge is the development of general schemes of learning that can effectively accommodate diverse model families, as data characteristics and task requirements can vary significantly depending on the problem at hand. In this context, we focus on decentralized federated learning of probabilistic generative classifiers, which model the joint probability distribution of the data in an efficient, flexible, and  interpretable manner,  making their application both feasible and valuable across a wide variety of classification tasks.

In the current work, we introduce a novel collaborative method designed for learning probabilistic generative classifiers over decentralized networks, enabling effective model training across distributed data owners. We argue that learning probabilistic generative classifiers in a serverless architecture is an important research line that has received limited attention due to the predominant use of deep neural network models in FL. This kind of model may not always be appropriate due to several factors, especially when the nature of the data deviates from typical large-scale local datasets and highly complex instances. Probabilistic models, in turn, emerge as an effective alternative when, for instance, we work with limited local datasets, deal with missing data, require model interpretability, need to model dependencies between variables, aim to incorporate prior knowledge, or when nodes have limited computing power, such as in edge devices or the Internet of Things \cite{GAMAZOREAL2023}. A representative case is the medical domain, where data scarcity and privacy constraints often make simpler and more interpretable models preferable \cite{Antunes22, JOCHEMS2017,Pezoulas2020}. Although FL is undergoing a very rapid development of new methods, only a few works have explored alternatives to neural networks \cite{NIPS:Smith17, lalitha2019decentralized, Navia22}, with most of them focusing on Bayesian models within a user-server architecture e.g. implementing federated variational inference \cite{boroujeni2025} or learning posterior distributions on the server side \cite{ashman2022}.

While other decentralized federated learning approaches have been proposed, they differ substantially in their assumptions and model scope. For instance, in \cite{Sun23}, a decentralized scheme based on undirected graphs is proposed to implement a federated version of gradient descent for empirical risk minimization. Although their method can theoretically be applied to any classifier learned via gradient-based optimization, their empirical validation is limited to neural networks trained on three datasets using relatively small communication networks (20 nodes) and large local training sets (thousands of samples). Gradient-based methods benefit from directly optimizing classification-specific objectives and can yield better classification performance in some cases. However, they lack many of the key benefits of probabilistic generative classifiers discussed earlier, such as interpretability, robustness to missing data, and the ability to incorporate prior knowledge.

On the other hand, gradient descent is not well-suited for learning probabilistic generative classifiers due to the structural constraints these models impose. In particular, as shown in \cite{perez2025}, risk-based calibration is more appropriate than gradient-based optimization for training probabilistic generative classifiers, which often require parameter constraints to be satisfied, e.g. semi-positive definite covariance matrices in quadratic discriminant analysis (QDA), or probability distributions that sum to one in naive Bayes (NB). Gradient descent may violate these constraints, requiring costly corrective projections that impair convergence and degrade classification performance.

In contrast, our work introduces a novel decentralized FL algorithm named Collaborative Risk Calibration (CRC) algorithm. This proposal is based on the risk-based calibration (RC) framework \cite{perez2025} and is designed to optimize the empirical risk when learning probabilistic generative classifiers—such as quadratic discriminant analysis or Bayesian network classifiers—across a network of users without requiring a central server. Unlike traditional Federated Averaging (FedAvg) schemes \cite{mcmahan17, Sun23}, which aggregate model parameters, CRC shares and aggregates local sufficient statistics to enable decentralized training.

We demonstrate that the CRC scheme effectively implements probabilistic models to address classification problems without the need for a central server. The analysis focuses on the model learning process, showing that local models converge to a common model that reduces the overall error of the network competitively compared to its centralized counterpart. This is validated through extensive experiments across a range of scenarios, including: i) diverse classification problems involving varying numbers of continuous and discrete variables, as well as different number of classes, ii) a range of graph structures providing communication networks of different topology and size, including dynamic connectivity, iii) non-identically distributed user datasets generated under various challenging conditions, iv) different sizes and degrees of fragmentation of local datasets, and v) different number of local iterations per round of learning.

The remainder of this paper is organized as follows. Section~\ref{sec:preliminaries} introduces probabilistic generative classifiers, the Risk-Based Calibration (RC) algorithm for learning this type of classifier, and the decentralized federated learning framework. Section~\ref{sec:CRC} presents Collaborative Risk-based Calibration (CRC), the novel federated decentralized version of RC. Section~\ref{sec:experiments} empirically evaluates CRC mainly in terms of the gap error between CRC and RC.  Finally, Section~\ref{sec:conclusions} summarizes the main contributions of this work. Additionally, the Appendix provides details on the naive Bayes (NB) classifiers and basic theoretical properties of CRC.

\section{Preliminaries}
\label{sec:preliminaries}

In this section, we introduce supervised learning of probabilistic generative classifiers and the risk-based calibration algorithm, which is adapted in this paper to learn probabilistic generative classifiers over decentralized architectures. We also introduces the framework used to conduct decentralized federated learning.

\subsection{Learning probabilistic generative classifiers} \label{sec:generative}

Supervised classification focuses on training a classifier from data to minimize the expected loss (or risk). A classifier $h \in \set{H}$ is a function that maps instances to labels, $h: \set{X} \rightarrow \set{Y}$, where  $\set{X} \subset \Real^q$ is the input space and $\set{Y}=\{1, \cdots, r\}$ the set of class labels. The loss, $l$, of a classifier $h$ at instance $(\x,y)$, with $\x \in \set{X}$ and $y \in \set{Y}$, is defined as $l: \set{H} \times (\set{X} \times \set{Y}) \rightarrow \Real^+$. Thus, supervised learning aims to find a classifier, $h \in \set{H}$, that minimizes the risk (expected loss):
\begin{equation}
\arg\min_{h} E_{p} [l(h,(\x,y))],
\end{equation}
where $p \in \Delta(\set{X} \times \set{Y})$ is the underlying joint distribution of the data.

In practice, the underlying joint distribution, $p$, is unknown, and we only have access to a supervised training set, $(X,Y)=\{(\x^i,y^i)\}_{i=1}^m$, with independent and identically distributed (i.i.d.) instances according to $p$. Generally, the 0-1 loss is considered, $l_{01}(h,(\x,y))= \1{y\neq h(\x)}$, and then, the objective is to minimize the empirical risk under the 0-1 loss or \emph{empirical error}:
\begin{equation}
\arg\min_{h} \frac{1}{m}\sum_{\x,y \in X,Y} l_{01}(h,(\x,y)).
\end{equation}

The learning problem is further simplified to be tractable. For instance, probabilistic generative classifiers solve the problem by approximating the underlying joint distribution using some specific parametric family of distributions $p_{\tht} \in \Delta(\set{X} \times \set{Y})$, where $\tht \in \Theta$ represent the parameters of the considered family of distributions. Usually, maximum likelihood or maximum a posteriori parameters are taken for $\tht$, and the classifier $h_{\tht}$ assigns to each $\x \in \set{X}$ the class label that maximizes the class conditional probability, $h_{\tht}(\x)=  \arg \max_y p_{\tht}(\x,y)$. When it is clear from the context, we denote the classifier $h_{\tht}$ simply as $h$. The intuition is that by having an accurate modeling of $p$ given by $p_{\tht}$, probabilistic generative classifiers can reach the minimum error, a.k.a. Bayes error. Typical examples of probabilistic generative classifiers include those based on joint probability distributions from the exponential family, such as the quadratic discriminant analysis and the NB classifiers from the family of classifiers based on Bayesian networks~\cite{bielza2014discrete,perez2006gaussian}.

By modeling the joint probability distribution, probabilistic generative classifiers offer several practical advantages over alternative approaches. This modeling enables intuitive interpretation of the parameters (e.g., means and covariances), facilitates the incorporation of prior knowledge via Bayesian frameworks (especially when using conjugate priors), and provides a comprehensive characterization of the data-generating process~\cite{augenstein2019generative,GM20}. Their relatively low number of parameters (e.g., NB with linear parameter growth) acts as a natural regularizer, helping to reduce overfitting and ensuring stable parameter updates. These models often generalize well even with small training sets, sometimes achieving optimal performance with sample sizes logarithmic in the number of parameters~\cite{ng2001discriminative}. Moreover, probabilistic generative models support synthetic data generation~\cite{Cuesta2019}, handle missing data robustly via the EM algorithm~\cite{dempster1977,Tolou23,Kim23}, and are well-suited to Bayesian decision theory, providing optimal predictions under cost-sensitive loss functions when class-conditional distributions are accurately modeled~\cite{Murphy12,elkan2001costsensitive}.

Probabilistic generative classifiers typically offer closed-form algorithms that estimate parameters in a single pass over the training data. These algorithms aim to maximize data-fitting objectives, such as the likelihood function. For certain parametric families of distributions, such as the exponential family, closed-form solutions based on data statistics exist that allow the computation of maximum likelihood or maximum a posteriori estimates given the data.
We formalize the closed-form algorithms \(a: \set{X}^m \times  \set{Y}^m \rightarrow \Real^o\) as the composition of a statistics mapping, 
\[s: \set{X}^m \times \set{Y}^m \rightarrow \Real^k,\]
and a parameter mapping 
\[ \theta: \Real^k \rightarrow \Real^o, \]
We assume that the statistics mapping is additive 
\[s(X,Y)=\sum_{x,y \in X,Y} s(x,y),\]
where $s(x,y): \set{X} \times \set{Y} \rightarrow \Real^k$. By letting $a= \theta \circ s$, the parameters obtained by a closed form algorithm are independent from data $X,Y$ given the statistics $s(X,Y)$. The computational complexity of a closed-form algorithm $a$ typically is $\mathcal{O}(m \times k + k \times o)$. For instance, the maximum likelihood parameters of NB can be obtained using a closed-form algorithm (see Appendix \ref{app:nb}).

Unfortunately, as probabilistic generative classifiers optimize likelihood-based objectives, they may obtain higher classification error than other kinds of classifiers that directly minimize a classification-specific loss \cite{jebara12}. For example, it is well established that the influence of the likelihood function on classification performance can diminish as the dimensionality of the feature space increases~\cite{friedman97}. The next section describes a recently proposed procedure that allows to learn the parameters of probabilistic generative classifiers by minimizing empirical risk, which improves classification performance while preserving the advantages of probabilistic generative classifiers.

\subsection{Risk-based calibration} \label{sec:rc}

The risk-based calibration (RC) algorithm~\cite{perez2025} is an iterative algorithm designed to reduce the empirical risk of a probabilistic generative classifier by leveraging an existing closed-form algorithm. RC progressively updates the statistics used to compute the parameter estimators guided by the soft 0-1 loss to find the parameters $\tht^*$ that minimize the corresponding empirical risk. The soft 0-1 loss is defined as:
$$l_{s01}(h,(\x,y))= 1- p_{\tht}(y|\x)= \sum_{y' \neq y}p_{\tht}(y'|\x),$$
where $p_{\tht}(y|\x)$ is obtained from the joint distribution by the Bayes rule. Intuitively, the soft 0-1 loss corresponds to the expected 0-1 loss of a randomized classifier that selects $y$ according to the class conditional distribution $p_{\tht}(y|\x)$.

\begin{algorithm}[t!]
\caption{Risk-based Calibration (RC)}
\label{alg:RC}
\begin{algorithmic}[1]
\STATE \textbf{Input:} Training set $(X,Y)$ with $m$ labeled instances; learning algorithm $a = \theta \circ s$, defined by the statistics mapping $s$ and the parameter mapping $\theta$; learning rate $lr$.
\STATE \textbf{Output:} Classifier $h$.
\STATE $\bd{s}^{0} \gets s(X,Y)$
\STATE $\tht^{0} \gets \theta(\bd{s}^{0})$
\STATE $t \gets 1$
\FOR{$t=1,2,\cdots, t^{max}$}
    \STATE $\bd{s}^{t} \gets \bd{s}^{t-1} + lr \cdot \big(s(X,Y) - s(X, \tht^{t-1})\big)$
    \STATE $\tht^{t} \gets \theta(\bd{s}^{t})$
\ENDFOR
\RETURN $h$ with parameters $\tht^{t^{max}}$
\end{algorithmic}
\end{algorithm}

Algorithm~\ref{alg:RC} shows the pseudo-code of RC. After initializing the statistics and the probabilistic generative classifier (steps 3 and 4), the algorithm updates the statistics and associated classifiers iteratively using local data and the classifier obtained in the previous iteration. The iterative calibration of the statistics continues until convergence, given in terms of the soft error (step 12). RC returns the classifier with the lowest training soft 0-1 loss (step 13). 

The intuition of the updating rule of RC is to increase $p_{\tht}(y|\x)$ and reduce $p_{\tht}(y'|\x)$ for each instance $(\x,y)$ by calibrating the set of statistics, $\bd{s}$, used to calculate the model parameters $\tht = \theta(\bd{s})$, according to the soft 0-1 loss. Assuming that $s(\x,y)$ denotes the statistics mapping from instance $(\x,y)$, the calibration of $\bd{s}$ is conducted by adding $s(\x,y)$ with weight $1-p_{\tht}(y|\x)$ and subtracting $s(\x,y')$ with weight $p_{\tht}(y'|\x)$ for all $y'\neq y$. That leads to the next update rule given the parameters $\theta(\bd{s})$ and the labeled instance $(\x,y)$:
\begin{equation}
\bd{s}= \bd{s} + s(x,y) - \sum_{y' \in \set{Y}} p_{\tht}(y'|\x)\cdot s(x,y').\label{eq:single.update}
\end{equation}
Given the additive nature of the statistics, $\bd{s}$, the RC update rule for the whole dataset $(X,Y)$ is given by step 8 in Algorithm \ref{alg:RC}, where $\tht^{t-1}$ denote the parameters at iteration $t-1$, i.e. $\theta(\bd{s}^{t-1})$. In this algorithm, with a slight abuse of notation, $s(X,Y)$ denotes the statistics mapping over the training set $(X,Y)$, while $s(X,\tht)$ represents the probabilistic statistics mapping according to the class conditional distribution $p_{\tht}(y|\x)$ defined as 
$$s(X,\tht)= \sum_{\x \in X} \sum_{y' \in \set{Y}} p_{\tht}(y'|\x)\cdot s(\x,y').$$

\subsection{Federated decentralized learning} \label{sec:DFL}

Decentralized FL is a collaborative, distributed machine learning approach designed to address classification problems across a network of users, without sharing private data or relying on a central server. In centralized FL, the nodes own the data and the server performs the aggregation of the user parameters to build the global model. In the absence of a server, the nodes both own the data and perform aggregation tasks, following a graph that defines the connections between them in the network.

The network is defined by an undirected graph $\set{G}=(\set{V}, \set{E})$ where $\set{V}= \{1,\ldots, n\}$ is the set of nodes and $\set{E} \subseteq \set{V} \times \set{V}$ is the set of edges that represent the existing connections between nodes. The set of neighbors of a node $v \in \set{V}$ is $\set{N}_v = \{u: (v,u) \in  \set{E}\}$. Each node of the network has each own local training set $(X_v, Y_v)$ with the corresponding number of instances $m_v$. We call global training data to the union of the local training datasets $(X,Y)=(\bigcup_{v \in \set{V}} X_v, \bigcup_{u \in \set{V}} Y_v)$

In general, the objective in federated learning \cite{Li20} is to find a classifier, $h$, that minimizes the weighted sum of the empirical risks over the local datasets:
\begin{equation}
\arg\min_{h} \hspace{0.1cm}  \sum_{v \in \set{V}}\frac{w_v}{m_v} \sum_{x,y \in X_v,Y_v} l_{01}(h,(x_v,y_v))), \label{eq:federated.error}
\end{equation}
where with a slight abuse in the notation $l_{01}(h,(X_v,Y_v)))$ is the empirical error of $h$ over the local training data of node $v$, and $w_v$ is the associated weight. The weight $w_v$ is commonly defined as $w_v=m_v/m$, giving equal importance to each data sample in the network, or as $w_v=1/n$, treating every node as equally important regardless of the number of samples \cite{marfoq23}. In this work, we focus on minimizing the soft 0-1 error over the global training set, corresponding to the federated error with weights $w_v = m_v/m$ for all $v \in \set{V}$ (Equation~\ref{eq:federated.error}).


\section{Collaborative risk-based calibration} \label{sec:CRC}

In this section, we present the novel federated decentralized version of RC called collaborative RC (CRC). 

\begin{algorithm}[t!]
\caption{Colaborative RC (CRC)}
\label{alg:CRC}
\begin{algorithmic}[1]
\STATE \textbf{Input:} Local training sets $(X_v,Y_v)$ with $m_v$ labeled instances for $v \in \set{V}$; Communication network given by the neighbors $\set{N}_v$ for $v \in \set{V}$; number of communication rounds, $t^{max}$; number of local iterations $iter$; equivalent sample size $m^0$; learning algorithm $a = \theta \circ s$, defined by the statistics mapping $s$ and the parameter mapping $\theta$.
\STATE \textbf{Output:} Local classifiers $h_v$ and local statistics $\bd{s}_v$ for $v \in \set{V}$ .

\FOR{$v \in \set{V}$}
\STATE $\bd{s}_v^{0} \gets initialize(m^0)$
\ENDFOR
\FOR{$t=1,2,\cdots, t^{max}$}
\FOR{$v \in \set{V}$}
\STATE $\bar{\bd{s}}_v^t \gets \frac{1}{|\set{N}_v|} \sum_{u \in \set{N}_v} \bd{s}_u^{t-1}$
\STATE $h_v^{t},\bd{s}_v^{t} \gets LocalRC(\bar{\bd{s}}_v^t, X_v, Y_v, iter, a)$  
\ENDFOR
\ENDFOR
\RETURN $h_v^{t^{max}}, \bd{s}_v^{t^{max}}$ for $v \in \set{V}$
\end{algorithmic}
\end{algorithm}

Algorithm~\ref{alg:CRC} shows the pseudo-code of CRC. The algorithm consists of a sequence of rounds in which the statistics of the neighborhood are aggregated by averaging them and then the statistics and its associated classifier are updated with local training data using LRC given in Algorithm \ref{alg:LRC}. The local statistics $\bd{s}_v^0$ are initialized so as they have an equivalent sample size of $m^0$, and the conditional distribution with parameters $\theta(\bd{s}_v^0)$ is uniform in $\Y$ for any $x \in \X$, for $v \in \set{V}$ (line 4, Algorithm~\ref{alg:CRC}). Thus, initially, local probabilistic generative models are the same model and they have no preference towards any class label. Appendix~\ref{app:nb} shows the details on the uniform initialization for NB. The aggregation rule of CRC is simply given by the average of the neighborhood statistics. All the local statistics share the same equivalent sample size $m^0$, and the updating rule of LRC not modify the equivalent sample size of the updated statistics. Thus the averaging aggregation maintain the equivalent sample size of the aggregated statistics. 

\begin{algorithm}[t]
\caption{Local RC (LRC)}
\label{alg:LRC}
\begin{algorithmic}[1]
\STATE \textbf{Input:} Aggregated statistics $\bar{\bd{s}}_v$; local training sets $(X_v,Y_v)$; number of iterations $iter$; learning algorithm $a = \theta \circ s$, defined by the statistics mapping $s$ and the parameter mapping $\theta$.
\STATE \textbf{Output:} Local classifiers $h_v$; Updated local statistics $\bd{s}_v$.

\STATE $\bd{s}_v \gets \bar{\bd{s}}_v$
\FOR{$i=1,2,\cdots ,iter$}
\STATE \hspace{1em} $\bd{s}_v \gets \bd{s}_v + s(X_v,Y_v) - s(X_v, \theta(\bd{s}_v))$
\ENDFOR
\RETURN $h_v$ with parameters $\theta(\bd{s}_v)$, $\bd{s}_v$
\end{algorithmic}
\end{algorithm}

Algorithm~\ref{alg:LRC} shows the pseudo-code of LRC. LRC is a simplified version of RC that updates input statistics using the RC updating rule a fixed number of times given by the parameter $iter$. The updating rule of LRC inherits the property of the updating rule of RC that maintain the sample size of the input statistics. The equivalent sample size of the input aggregated statistics, $m^0$, represents their inertia to the local updates of LRC. It plays the role of a local learning rate that depends on the sample size of the local data $m_v$ and is given by $m_v/m^0$. Thus, the updates of a user with more local data can have a stronger impact in the generated local statistics. 

\begin{figure}[ht]
    \centering
    \includegraphics[width=\linewidth]{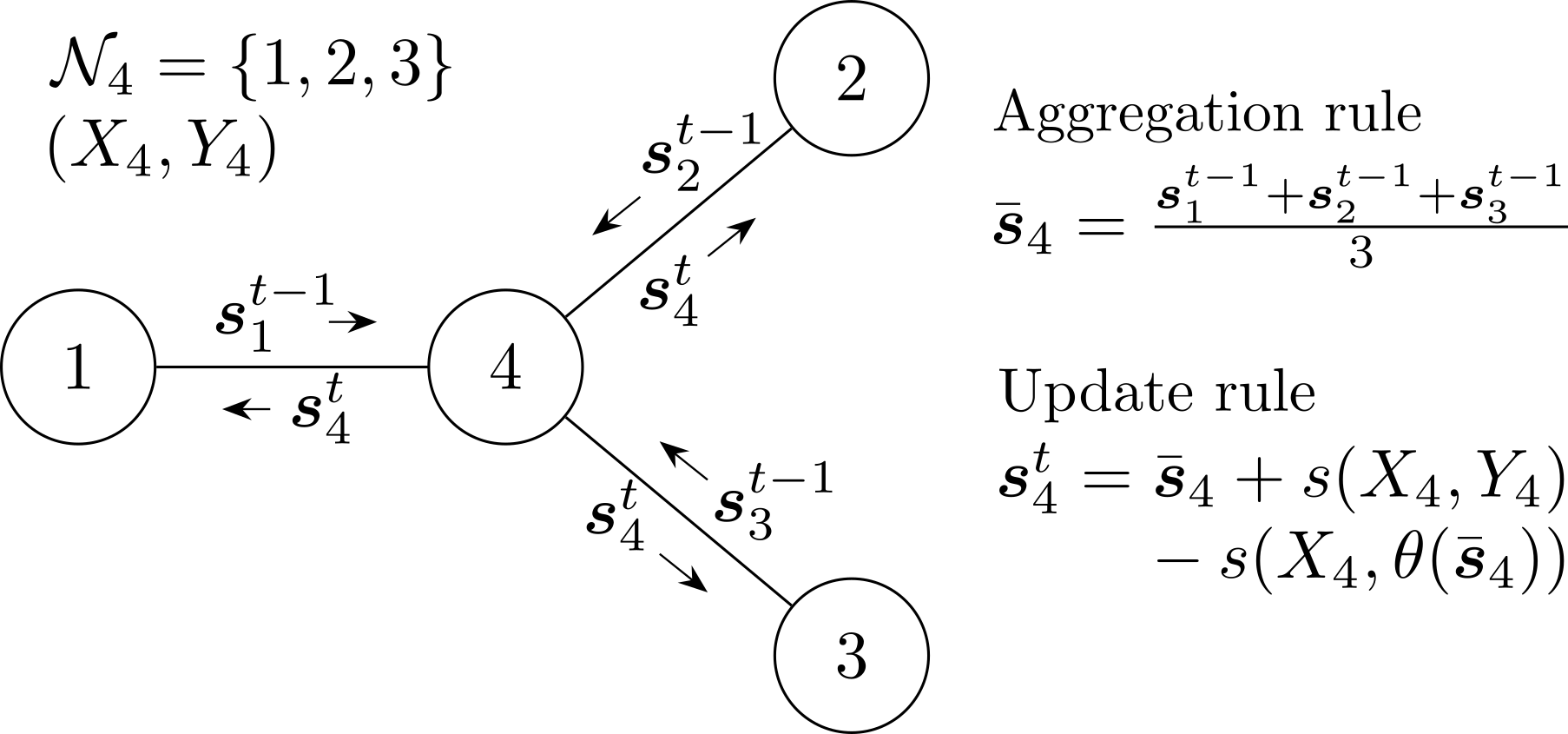}
    \caption{This figure shows an example of round $t$ of CRC with a single local update of the statistics for the node $4$ using the local data $(X_4,Y_4)$ and the communications with its neighbors, $\set{N}_4=\{1,2,3\}$. The update is given by the aggregation rule and the update rule.}
    \label{fig:CRC}
\end{figure}

In line with the decentralized nature of CRC, Figure~\ref{fig:CRC} shows an example of the round $t$ of CRC for the user $4$ with neighborhood $\set{N}_4=\{1,2,3\}$. At each round, the behavior of a user is independent from the rest of the network, given its neighbors. First, node $4$ receives the statistics of $\set{N}_4$ produced at previous round, $\{\bd{s}_1^{t-1}, \bd{s}_2^{t-1}, \bd{s}_3^{t-1}\}$, then node $4$ combines the neighbor statistics to get $\bar{\bd{s}}_4^t$ using the aggregation rule (line 4, Algorithm \ref{alg:LRC}) to get $\bar{\bd{s}}_4^t$. Next, $4$ updates the aggregated statistics using local data using the update rule (line 8 in Algorithm \ref{alg:CRC}) to get $\bd{s}_4^{t}$, and it communicates the updated statistics to its neighbors $\set{N}_4$.

\subsection{On the equivalence of CRC and RC}

In despite of its extreme simplicity, CRC is able to replicate RC by considering a full graph as the communication network without further assumptions on the available local training data.

\begin{theorem}\label{thm:equivalence}
Let $(\set{V},\set{E})$ be a communication network with $\set{N}_v = \set{V}$ for all $v \in \set{V}$. Let $\{(X_v,Y_v): v \in \set{V}\}$ be a partition of the global training data $(X,Y)$, where $m = \sum_{v\in \set{V}} m_v$. Consider RC with learning rate $lr > 0$, using the global training data $(X,Y)$ with uniform initialization and a single iteration, i.e., $iter = 1$. Now consider CRC with equivalent sample size $m^0 = \frac{m}{lr \cdot n}$, using the local data $(X_v,Y_v)$ for each $v \in \set{V}$ with uniform initialization and one iteration ($iter = 1$). Then, all local parameters in the aggregation step of CRC coincide with the parameters of RC at the previous iteration:
$$\theta(\bar{\bd{s}}^{t}_v) = \theta(\bd{s}^{t-1}_g), \quad \text{for } t = 1, \dots, t^{\max}, \text{ and } v \in \set{V}.$$
\end{theorem}

This result shows that in the extreme case of having a complete communication network, CRC can replicate the behavior of RC. In most practical scenarios, a complete communication network lacks of interest and sparse communication networks are available. However, in general, as the communication network is more dense (closer to the complete subgraph) the sequence of local classifiers obtained by CRC is closer to that obtained by RC, and at the same time, local classifiers are more similar between them. The sparsest communication network with a single connected component is a tree, which has only $n-1$ edges. In the experiments, we empirically show that the gap between CRC and RC is smaller than $0.01$ for most of the datasets. We also show that the differences become smaller just by adding an small percentage of the possible edges. 

Theorem \ref{thm:equivalence} does not require assumptions about the local data generation process and thus the equivalence between CRC using full communication network and RC holds non-i.i.d. local training sets. In the experiments, we show that this holds even for sparsest communication networks and that by adding an small percentage of the possible edges CRC can obtain gaps smaller than $0.01$ for most of the datasets.

Using Theorem \ref{thm:equivalence}, we propose the next heuristic to select $m^0$ for a communication network given a learning rate of RC $lr$:
\begin{equation}
m^0=\frac{m}{lr\cdot n},
\end{equation}
where $m$ is the size of the global training data in the network. For instance, for $lr=0.05$ we have $m^0=\frac{20*m}{n}$, and for the particular case of constant local training set size $m^0= 20 \cdot m_v$.

\section{Experimental results}
\label{sec:experiments}

This section presents experimental results that show the effectiveness of CRC to learn probabilistic generative classifiers in federated decentralized settings. More specifically, we evaluate the robustness, scalability, and convergence behavior of CRC under a broad variety of configurations for the communication network and local training sets relevant to real-world federated decentralized scenarios. 

All experiments were conducted using the NB classifier with both discrete and continuous variables, trained via a closed-form maximum likelihood algorithm. The objective is to evaluate the collaborative risk-based calibration (\textbf{CRC}) learning algorithm under this same training procedure. Appendix \ref{app:nb} provides the details on the NB classifier and the maximum likelihood closed-form learning algorithm. CRC learns a model for each user in a decentralized network of size $n$, where each user has a local training set of size $m_v$. CRC will be compared against the risk-based calibration algorithm (\textbf{RC}) using the same closed-form algorithm. RC learns NB by using the global training set given by the union of the local training data from all nodes in the network. NB learned using RC represents the \textbf{gold-standard} since is a centralized version of CRC with the entire training data in the network, and it provides and upper bound for the performance for the local NB models learned using CRC. We also include, as a reference, the results obtained by NB with the maximum likelihood parameters using the global training set (\textbf{ML}).

\textbf{Training} and \textbf{test errors} are the average 0-1 loss of NB in the global training data and a disjoint test data, respectively. In the case of CRC, we have $n$ different NB classifiers, one for each user, and for CRC the training and test errors corresponds to the average training and test errors of the local NB classifiers across the different users. We would like to highlight that training error for CRC is the average of NB classifiers calibrated using only local training sets but tested on the global training set. We call the \textbf{training} and \textbf{test} \textbf{gaps} between CRC and RC the differences between their training and test errors, respectively. We are primarily interested in the training and test gaps between CRC and RC, since the errors of CR are lower bounds to those of CRC. All the experiments have been repeated 5 times using different splits for the global training and test sets, and randomly generated communication networks. All the scores that we report in the tables are averaged across the 5 repetitions. In the tables summarizing the results, we highlight gaps in $(0.05,0.01]$ in \textbf{\textcolor{gray}{bold gray}} and those lower than $0.01$ in \textbf{bold black}. We consider that CRC has converged to RC when the gaps are smaller than $0.01$.

\begin{figure}[t]
\centering
\includegraphics[width=\linewidth]{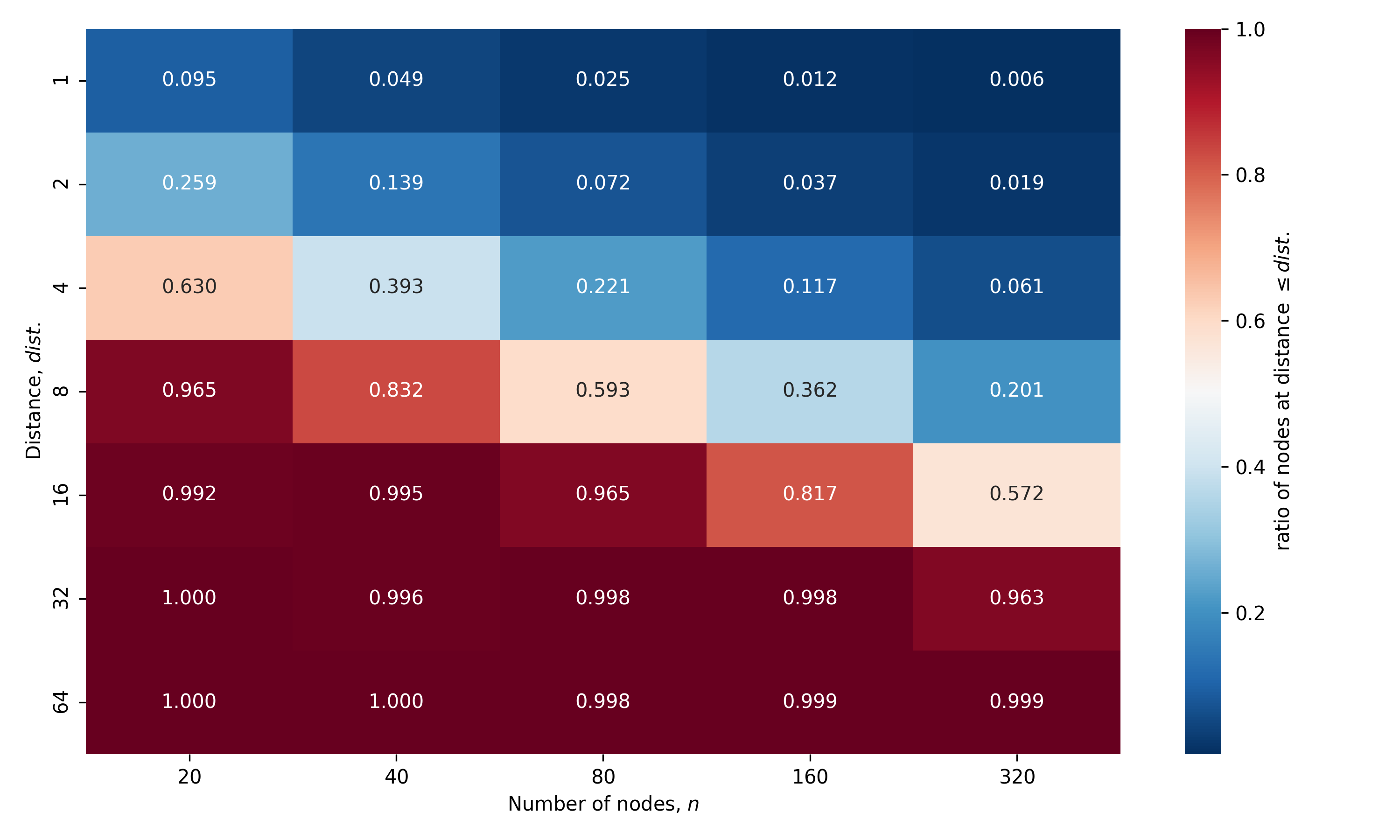}
\caption{Average ratio of nodes at a distance less than or equal to a given value relative to $n$, in random trees. This ratio indicates the fraction of nodes that can influence a given node and relates to the difficulty of convergence of CRC.}
\label{fig:distance.heatmap}
\end{figure}

The default parameters of the experiments are: number of nodes $n=50$, size of the local training datasets $m_v=50$ (i.i.d.), number of maximum communication rounds $t^{max}=64$, number of LRC iterations $iter=1$, learning rate of RC with global data $lr=0.05$, and equivalent sample size $m^0=m_v/lr$. We show the results of CRC and RC for round $t^{max}=64$. The default communication topology is a tree, generated at random. The trees correspond to the sparsest connected graphs, with only $(n-1)$ edges compared to the possible $n(n-1)/2$. We have selected the sparsest connected graphs to expose CRC to the most challenging communication network topology. As $n$ increases the sparseness of the tree increases, being the \textbf{sparseness} the ratio of the possible edges included, given by $2/(n-1)$. 
Figure~\ref{fig:distance.heatmap} shows the effect of tree size $n$ on the average proportion of nodes within a given communication distance, which highlights the increase in the difficulty of the cooperative learning as the size of the tree increases. To further increase the difficulty of learning accurate local NB classifiers using CRC, we have selected a relatively high default size of the communication network ($n=50$) and a small default size of the local training sets ($m_v=50$) \cite{Sun23}.

Experiments will be conducted on 16 datasets. To allow experiments with large communication networks, we have selected datasets with a minimum of 16000 samples. Continuous variables with ten or fewer distinct values have been treated as discrete variables. Table~\ref{tab:datasets} summarizes the main features of the datasets. Table~\ref{tab:effect_train_size_global_gap} shows the test gaps between RC using different training set sizes $m \in \{20,40,80,160,320,640,1280,2650,5120\}$ and the RC trained using $m=10240$ as the gold-standard. These gaps indicate the amount of training data required to obtain errors close to the RC using the largest training set, which is the most deterministic factor in federated decentralized learning. For instance, given the default local training size $m_v=50$. The last column in Table~\ref{tab:effect_train_size_global_gap} shows the test error of the gold standard for reference. Table~\ref{tab:datasets} shows the datasets ordered using the gaps of Table~\ref{tab:effect_train_size_global_gap} to ease the analysis of the experimental results. Note that the datasets at the bottom of Table~\ref{tab:effect_train_size_global_gap} correspond to more complex classification problems. For example, the Letter dataset involves predicting 26 distinct class labels. Consequently, small local datasets are not sufficient to train accurate classifiers. Even with a training set of size 1280, the gap relative to the reference remains substantial (0.13).

\begin{table}
\caption{Dataset information. Columns: \texttt{name} the names of the datasets, $m$ the number of instances, $d$ the number of features, $disc.$ the number of discrete features, $cont.$ the number of continuous features, and $|\mathcal{Y}|$ the number of class labels.}\label{tab:datasets}
\begin{tabular}{l|rrrrr}
\toprule
name & $m$ & $d$ & $disc.$ & $cont.$ & $|\mathcal{Y}|$ \\
\midrule
skin & 245056 & 3 & 0 & 3 & 2 \\
pulsar stars & 17897 & 8 & 0 & 8 & 2 \\
catsvsdogs & 23262 & 512 & 0 & 512 & 2 \\
bank marketing & 45211 & 16 & 7 & 9 & 2 \\
default credit card & 29999 & 23 & 5 & 18 & 2 \\
soft heart disease & 319795 & 17 & 12 & 5 & 2 \\
adult & 48841 & 14 & 5 & 9 & 2 \\
run or walk & 88588 & 6 & 0 & 6 & 2 \\
smartgrid stability & 59999 & 13 & 0 & 13 & 2 \\
smoking & 55691 & 24 & 6 & 18 & 2 \\
yearbook & 37921 & 512 & 0 & 512 & 2 \\
secondary mushroom & 61067 & 19 & 12 & 7 & 2 \\
fashion mnist & 70000 & 512 & 0 & 512 & 10 \\
cifar10 & 60000 & 512 & 0 & 512 & 10 \\
mnist & 70000 & 510 & 0 & 510 & 10 \\
letter recognition & 20000 & 16 & 0 & 16 & 26 \\
\bottomrule
\end{tabular}
\end{table}

\begin{table*}[t]
\centering
\caption{Test gaps between RC for different global training set sizes, $m$, and RC with $m=10240$. The last column shows the test error of RC for $m=10240$. Gaps $< 0.01$ are bold black; gaps $< 0.05$ are bold gray.
}
\label{tab:effect_train_size_global_gap}
\begin{tabular}{l|ccccccccc|c}
\toprule
Dataset & 20 & 40 & 80 & 160 & 320 & 640 & 1280 & 2560 & 5120 & 10240 \\
\midrule
skin & \textbf{\textcolor{gray}{0.01}} & \textbf{\textcolor{gray}{0.01}} & \textbf{0.00} & \textbf{0.00} & \textbf{0.00} & \textbf{0.00} & \textbf{0.00} & \textbf{0.00} & \textbf{0.00} & 0.07 \\
pulsar & \textbf{\textcolor{gray}{0.02}} & \textbf{\textcolor{gray}{0.01}} & \textbf{0.01} & \textbf{0.00} & \textbf{0.00} & \textbf{0.00} & \textbf{0.00} & \textbf{0.00} & \textbf{0.00} & 0.03 \\
catsvsdogs & \textbf{\textcolor{gray}{0.03}} & \textbf{\textcolor{gray}{0.02}} & \textbf{\textcolor{gray}{0.02}} & \textbf{\textcolor{gray}{0.01}} & \textbf{0.01} & \textbf{0.01} & \textbf{0.00} & \textbf{0.00} & \textbf{0.00} & 0.01 \\
bank & 0.05 & \textbf{\textcolor{gray}{0.03}} & \textbf{\textcolor{gray}{0.01}} & \textbf{\textcolor{gray}{0.01}} & \textbf{0.00} & \textbf{0.00} & \textbf{0.00} & \textbf{0.00} & \textbf{0.00} & 0.11 \\
default & 0.07 & 0.07 & \textbf{\textcolor{gray}{0.05}} & \textbf{\textcolor{gray}{0.03}} & \textbf{\textcolor{gray}{0.02}} & \textbf{0.00} & \textbf{0.00} & \textbf{0.00} & \textbf{0.00} & 0.18 \\
soft & 0.09 & 0.06 & \textbf{\textcolor{gray}{0.03}} & \textbf{0.01} & \textbf{0.00} & \textbf{0.00} & \textbf{0.00} & \textbf{0.00} & \textbf{0.00} & 0.10 \\
adult & 0.09 & 0.08 & \textbf{\textcolor{gray}{0.05}} & \textbf{\textcolor{gray}{0.04}} & \textbf{\textcolor{gray}{0.02}} & \textbf{0.01} & \textbf{0.00} & \textbf{0.00} & \textbf{0.00} & 0.16 \\
run & 0.11 & 0.05 & 0.05 & \textbf{\textcolor{gray}{0.03}} & \textbf{\textcolor{gray}{0.02}} & \textbf{0.00} & \textbf{0.00} & \textbf{0.00} & \textbf{0.00} & 0.18 \\
smartgrid & 0.14 & 0.07 & \textbf{\textcolor{gray}{0.04}} & \textbf{\textcolor{gray}{0.02}} & \textbf{0.01} & \textbf{0.01} & \textbf{0.00} & \textbf{0.00} & \textbf{0.00} & 0.02 \\
smoking & 0.11 & 0.09 & 0.07 & \textbf{\textcolor{gray}{0.05}} & \textbf{\textcolor{gray}{0.03}} & \textbf{\textcolor{gray}{0.02}} & \textbf{\textcolor{gray}{0.01}} & \textbf{0.01} & \textbf{0.00} & 0.26 \\
yearbook & 0.14 & 0.10 & 0.07 & 0.05 & \textbf{\textcolor{gray}{0.05}} & \textbf{\textcolor{gray}{0.04}} & \textbf{\textcolor{gray}{0.03}} & \textbf{\textcolor{gray}{0.02}} & \textbf{0.01} & 0.08 \\
secondary & 0.18 & 0.16 & 0.11 & 0.07 & \textbf{\textcolor{gray}{0.04}} & \textbf{\textcolor{gray}{0.02}} & \textbf{0.01} & \textbf{0.00} & \textbf{0.00} & 0.22 \\
fashion & 0.36 & 0.26 & 0.18 & 0.13 & 0.08 & 0.06 & \textbf{\textcolor{gray}{0.04}} & \textbf{\textcolor{gray}{0.03}} & \textbf{0.01} & 0.14 \\
cifar10 & 0.42 & 0.33 & 0.22 & 0.15 & 0.11 & 0.08 & 0.05 & \textbf{\textcolor{gray}{0.03}} & \textbf{\textcolor{gray}{0.02}} & 0.15 \\
mnist & 0.49 & 0.38 & 0.27 & 0.18 & 0.10 & 0.06 & \textbf{\textcolor{gray}{0.04}} & \textbf{\textcolor{gray}{0.02}} & \textbf{0.01} & 0.04 \\
letter & 0.77 & 0.77 & 0.73 & 0.44 & 0.33 & 0.22 & 0.13 & 0.06 & \textbf{\textcolor{gray}{0.02}} & 0.19 \\
\bottomrule
\end{tabular}
\end{table*}

\textbf{Organization}: To provide a comprehensive and systematic evaluation of CRC, the experiments have been divided into subsections covering the following fundamental aspects in federated decentralized learning. We first analyze the convergence of local models by examining how their errors evolve during training under our default experimental conditions. Next, we examine the robustness of CRC to different communication network topologies ($\set{N}_v$) by using different undirected graphs. This is motivated by the critical role that the network plays in information propagation and convergence speed. Then, we address robustness to non-i.i.d. local datasets, reflecting the practical challenge of heterogeneous data distributions across clients. We also explore the effect of the size of the local dataset ($m_v$) and the effect of the size of the communication network ($n$), as both influence efficiency and scalability. In addition, we analyze the effect of data fragmentation with constant global training data ($m= n \cdot m_v$), to better understand how splitting the same amount of data among more clients affects the learning process. The impact of the number of local iterations per round ($iter$) is also investigated, given its importance in the trade-off between local computation and communication frequency. Finally, we evaluate the robustness to dynamic changes in the communication network topology ($\set{N}_v^t$), which simulates real-world scenarios where connections may vary over time.

\subsection{Convergence of local models}\label{sec:exper.convergence}

In federated learning, convergence of decentralized models to the global centralized optimum is a fundamental property. Understanding how quickly local models trained with CRC approach the gold standard RC model is key to validating its efficiency.

The experiments have been carried out with the default parameter values. Table~\ref{tab:convergence_summary_gap} shows the training and test gaps between ML and between CRC and RC. In addition, we include the standard deviation for training and test errors of local NB classifiers across the different users (std.). The standard deviation measures the similarity of local models trained with CRC, while the gap measures the convergence to RC. The table also includes the training and test errors for RC as a reference. 

CRC outperforms ML (trained with $2500$ samples) in 15 out of 16 datasets in terms of both training and test errors. RC shows similar training and test errors, and CRC has the same train and test gaps, which highlight the robustness of CRC to overfitting. CRC has training and test gaps smaller than $0.01$ ($0.05$) in 10 (13) and 9 (14) out of 16 datasets, respectively. In addition, the standard deviations of training and test errors of local models trained with CRC are smaller than $0.01$ ($0.05$) in $13$ ($16$) out of $16$ datasets. These results clearly show that the local models obtained by CRC have converged to a similar model that is close to the global model. 

\begin{table}[h]
\centering
\caption{Convergence of local models: training and test gaps at round $t = 64$ between ML and between CRC and RC, respectively. The column std. shows the standard deviations for the errors obtained by the local NB classifiers learned using CRC. 
}
\label{tab:convergence_summary_gap}
\begin{tabular}{l|c|cc|c|c|cc|c}
\toprule
 & \multicolumn{4}{c|}{Training} & \multicolumn{4}{c}{Test} \\
 & \multicolumn{1}{c|}{ML} & \multicolumn{2}{c|}{CRC} & \multicolumn{1}{c|}{RC} & \multicolumn{1}{c|}{ML} & \multicolumn{2}{c|}{CRC} & \multicolumn{1}{c}{RC} \\
Dataset & gap & gap & std. & error & gap & gap & std. & error \\
\midrule
skin & \textbf{0.01} & \textbf{0.00} & \textbf{0.00} & 0.07 & \textbf{0.01} & \textbf{0.00} & \textbf{0.00} & 0.07 \\
pulsar & \textbf{\textcolor{gray}{0.02}} & \textbf{0.00} & \textbf{0.00} & 0.02 & \textbf{\textcolor{gray}{0.03}} & \textbf{0.00} & \textbf{0.00} & 0.03 \\
catsvsdogs & \textbf{\textcolor{gray}{0.04}} & \textbf{0.01} & \textbf{0.00} & 0.00 & \textbf{\textcolor{gray}{0.02}} & \textbf{0.00} & \textbf{0.00} & 0.01 \\
bank & \textbf{\textcolor{gray}{0.03}} & \textbf{0.00} & \textbf{0.00} & 0.10 & \textbf{\textcolor{gray}{0.03}} & \textbf{0.00} & \textbf{0.00} & 0.11 \\
default & 0.19 & \textbf{0.01} & \textbf{0.01} & 0.18 & 0.19 & \textbf{0.01} & \textbf{0.01} & 0.18 \\
soft & \textbf{\textcolor{gray}{0.03}} & \textbf{0.00} & \textbf{0.00} & 0.10 & \textbf{\textcolor{gray}{0.03}} & \textbf{0.00} & \textbf{0.00} & 0.10 \\
adult & 0.06 & \textbf{0.01} & \textbf{0.00} & 0.15 & \textbf{\textcolor{gray}{0.05}} & \textbf{0.01} & \textbf{0.00} & 0.16 \\
run & \textbf{\textcolor{gray}{0.04}} & \textbf{0.00} & \textbf{0.01} & 0.19 & \textbf{\textcolor{gray}{0.04}} & \textbf{0.00} & \textbf{0.01} & 0.19 \\
smartgrid & \textbf{\textcolor{gray}{0.01}} & \textbf{0.00} & \textbf{0.00} & 0.02 & \textbf{\textcolor{gray}{0.01}} & \textbf{0.00} & \textbf{0.00} & 0.02 \\
smoking & 0.07 & \textbf{\textcolor{gray}{0.01}} & \textbf{0.01} & 0.24 & \textbf{\textcolor{gray}{0.05}} & \textbf{\textcolor{gray}{0.01}} & \textbf{0.01} & 0.27 \\
yearbook & 0.15 & \textbf{\textcolor{gray}{0.03}} & \textbf{0.01} & 0.04 & 0.08 & \textbf{0.01} & \textbf{0.01} & 0.10 \\
secondary & 0.15 & \textbf{\textcolor{gray}{0.02}} & \textbf{\textcolor{gray}{0.01}} & 0.21 & 0.14 & \textbf{\textcolor{gray}{0.02}} & \textbf{\textcolor{gray}{0.01}} & 0.22 \\
fashion & 0.38 & 0.21 & \textbf{\textcolor{gray}{0.02}} & 0.00 & 0.24 & 0.12 & \textbf{\textcolor{gray}{0.01}} & 0.17 \\
cifar10 & 0.42 & 0.15 & \textbf{0.01} & 0.00 & 0.26 & 0.08 & \textbf{0.00} & 0.19 \\
mnist & 0.40 & 0.34 & \textbf{\textcolor{gray}{0.04}} & 0.00 & 0.35 & 0.31 & \textbf{\textcolor{gray}{0.03}} & 0.07 \\
letter & 0.50 & 0.75 & \textbf{0.00} & 0.06 & 0.38 & 0.58 & \textbf{0.00} & 0.26 \\
\bottomrule
\end{tabular}
\end{table}

Due to lack of overfitting, from here on, we will focus the analysis of CRC in terms of test gap with respect to the gold-standard represented by the RC using the global training data.

\subsection{Sparse communication networks}
\label{sec:exper.topology}

In this section, we evaluate the robustness of CRC to different communication network topologies, which strongly affect the learning performance in decentralized federated learning. The communication graph structure determines how efficiently information propagates across the network. Sparse or poorly connected networks may slow convergence or even prevent consensus, while well-connected graphs facilitate faster and more stable convergence. In fact, Theorem~\ref{thm:equivalence} proves that CRC obtain local models equivalent to the RC global model using full communication network. Thus, this section focuses on analyzing empirically the behavior of CRC for sparse communication networks.

In this experiment, we vary the communication topology while keeping the rest of default parameters. Specifically, we depart from the sparsest possible communication network given by a tree with only $4\%$ of the possible edges, and increase connectivity by adding 10 ($5\%$), 20 ($6\%$), 40 ($7\%)$, and 80 ($11\%$) random edges among the possible $1225$ edges for $n=50$. These graphs represent randomly generated sparse connected communication networks with a single connected component. We also include the chain structure representing a particular tree with the worst communication properties.

Table~\ref{tab:topology_gap_iid} summarizes the test gaps between CRC for different network topologies and RC using global training set. CRC using tree structure obtain gaps smaller than $0.01$ ($0.05$) in $9$ ($15$) out of $16$ datasets and using chains can obtain gaps close to the ones obtained with trees. Besides, CRC using trees with $10$ randomly added edges in $13$ ($15$) out of $16$. These results highlight the robustness of CRC using sparse communication networks and the benefits by adding a very small quantity of edges.

\begin{table}[h]
\centering
\caption{Influence of communication network sparsity: test gaps between CRC using different sparse communication network topologies and RC at round $t=64$.
}
\label{tab:topology_gap_iid}
\begin{tabular}{l|cccccc}
\toprule
\multicolumn{1}{c|}{} & \multicolumn{6}{c}{Topology of $\set{N}_l$} \\
Dataset & tree & chain & +10 & +20 & +40 & +80 \\
\midrule
skin & \textbf{0.00} & \textbf{0.00} & \textbf{0.00} & \textbf{0.00} & \textbf{0.00} & \textbf{0.00} \\
pulsar & \textbf{0.00} & \textbf{0.00} & \textbf{0.00} & \textbf{0.00} & \textbf{0.00} & \textbf{0.00} \\
catsvsdogs & \textbf{0.00} & \textbf{0.00} & \textbf{0.00} & \textbf{0.00} & \textbf{0.00} & \textbf{0.00} \\
bank & \textbf{0.00} & \textbf{0.00} & \textbf{0.00} & \textbf{0.00} & \textbf{0.00} & \textbf{0.00} \\
default & \textbf{0.01} & \textbf{0.01} & \textbf{0.01} & \textbf{0.00} & \textbf{0.00} & \textbf{0.00} \\
soft & \textbf{0.00} & \textbf{0.00} & \textbf{0.00} & \textbf{0.00} & \textbf{0.00} & \textbf{0.00} \\
adult & \textbf{0.01} & \textbf{\textcolor{gray}{0.01}} & \textbf{0.01} & \textbf{0.00} & \textbf{0.00} & \textbf{0.00} \\
run & \textbf{0.01} & \textbf{0.01} & \textbf{0.00} & \textbf{0.00} & \textbf{0.00} & \textbf{0.00} \\
smartgrid & \textbf{0.00} & \textbf{0.00} & \textbf{0.00} & \textbf{0.00} & \textbf{0.00} & \textbf{0.00} \\
smoking & \textbf{\textcolor{gray}{0.02}} & \textbf{\textcolor{gray}{0.02}} & \textbf{\textcolor{gray}{0.01}} & \textbf{0.01} & \textbf{0.01} & \textbf{0.00} \\
yearbook & \textbf{\textcolor{gray}{0.01}} & \textbf{\textcolor{gray}{0.02}} & \textbf{0.01} & \textbf{0.00} & \textbf{0.00} & \textbf{0.00} \\
secondary & \textbf{\textcolor{gray}{0.03}} & \textbf{\textcolor{gray}{0.03}} & \textbf{\textcolor{gray}{0.02}} & \textbf{\textcolor{gray}{0.01}} & \textbf{0.01} & \textbf{0.01} \\
fashion & \textbf{\textcolor{gray}{0.02}} & \textbf{\textcolor{gray}{0.03}} & \textbf{0.01} & \textbf{0.01} & \textbf{0.00} & \textbf{0.00} \\
cifar10 & \textbf{\textcolor{gray}{0.02}} & \textbf{\textcolor{gray}{0.03}} & \textbf{0.01} & \textbf{0.01} & \textbf{0.00} & \textbf{\textcolor{gray}{0.01}} \\
mnist & \textbf{\textcolor{gray}{0.05}} & \textbf{\textcolor{gray}{0.03}} & \textbf{\textcolor{gray}{0.01}} & \textbf{0.01} & \textbf{0.00} & \textbf{0.00} \\
letter & 0.46 & 0.37 & 0.46 & 0.49 & 0.51 & 0.52 \\
\bottomrule
\end{tabular}
\end{table}

\subsection{Non-i.i.d. data}
\label{sec:exper.iid}

Real-world federated data is rarely i.i.d., which can severely impact decentralized learning. Thus, testing CRC under extreme non-i.i.d. scenarios is critical to assess its robustness.

We generate three types of particularly extreme non-i.i.d. local datasets: (i) drift in $p(\x)$, by splitting sorted data according to the first PCA component; (ii) drift in $p(y)$, by assigning each user samples from a single class; (iii) drift in $p(\x,y)$, combining input and label drifts. The local datasets have been generated in such a way that global training set remains i.i.d. and, thus, RC learns the NB classifier using i.i.d. training data. According to Theorem~\ref{thm:equivalence} CRC is able to obtain the same models as RC with the global training set with the full communication network. Thus, in these experiments we also vary the communication topology departing from trees and adding an small amount of edges, while keeping the rest of default parameters.

\begin{table}[h]
\centering
\caption{Influence of non-i.i.d data with extreme drifts in $p(\x)$: test gaps between CRC using different sparse communication network topologies and RC at round $t=64$. 
}
\label{tab:topology_gap_X}
\begin{tabular}{l|ccccc}
\toprule
\multicolumn{1}{c|}{} & \multicolumn{5}{c}{Topology of $\set{N}_l$} \\
Dataset & tree & +10 & +20 & +40 & +80 \\
\midrule
skin & \textbf{0.00} & \textbf{0.00} & \textbf{0.00} & \textbf{0.00} & \textbf{0.00} \\
pulsar & \textbf{\textcolor{gray}{0.03}} & \textbf{\textcolor{gray}{0.02}} & \textbf{\textcolor{gray}{0.01}} & \textbf{0.01} & \textbf{0.00} \\
catsvsdogs & \textbf{0.01} & \textbf{0.00} & \textbf{0.00} & \textbf{0.00} & \textbf{0.00} \\
bank & \textbf{0.01} & \textbf{0.00} & \textbf{0.00} & \textbf{0.00} & \textbf{0.00} \\
default & \textbf{\textcolor{gray}{0.02}} & \textbf{\textcolor{gray}{0.02}} & \textbf{\textcolor{gray}{0.01}} & \textbf{0.01} & \textbf{0.00} \\
soft & \textbf{0.00} & \textbf{0.00} & \textbf{0.00} & \textbf{0.00} & \textbf{0.00} \\
adult & \textbf{\textcolor{gray}{0.03}} & \textbf{\textcolor{gray}{0.02}} & \textbf{\textcolor{gray}{0.01}} & \textbf{0.01} & \textbf{0.00} \\
run & \textbf{\textcolor{gray}{0.03}} & \textbf{\textcolor{gray}{0.01}} & \textbf{0.01} & \textbf{0.00} & \textbf{0.00} \\
smartgrid & \textbf{0.00} & \textbf{0.00} & \textbf{0.00} & \textbf{0.00} & \textbf{0.00} \\
smoking & \textbf{\textcolor{gray}{0.02}} & \textbf{\textcolor{gray}{0.02}} & \textbf{\textcolor{gray}{0.01}} & \textbf{0.01} & \textbf{0.01} \\
yearbook & \textbf{\textcolor{gray}{0.02}} & \textbf{\textcolor{gray}{0.01}} & \textbf{\textcolor{gray}{0.01}} & \textbf{\textcolor{gray}{0.01}} & \textbf{0.01} \\
secondary & 0.07 & \textbf{\textcolor{gray}{0.05}} & \textbf{\textcolor{gray}{0.04}} & \textbf{\textcolor{gray}{0.03}} & \textbf{\textcolor{gray}{0.02}} \\
fashion & \textbf{\textcolor{gray}{0.03}} & \textbf{\textcolor{gray}{0.02}} & \textbf{0.01} & \textbf{0.01} & \textbf{0.00} \\
cifar10 & \textbf{\textcolor{gray}{0.02}} & \textbf{0.01} & \textbf{0.01} & \textbf{0.01} & \textbf{0.00} \\
mnist & 0.06 & \textbf{\textcolor{gray}{0.01}} & \textbf{0.01} & \textbf{0.01} & \textbf{0.00} \\
letter & 0.47 & 0.56 & 0.61 & 0.63 & 0.64 \\
\bottomrule
\end{tabular}
\end{table}
\begin{table}[h]
\centering
\caption{Influence of non-i.i.d data with extreme drifts in $p(y)$: test gaps between CRC using different sparse communication network topologies and RC at round $t=64$. 
}
\label{tab:topology_gap_Y}
\begin{tabular}{l|ccccc}
\toprule
\multicolumn{1}{c|}{} & \multicolumn{5}{c}{Topology of $\set{N}_l$} \\
Dataset & tree & +10 & +20 & +40 & +80 \\
\midrule
skin & \textbf{\textcolor{gray}{0.01}} & \textbf{0.01} & \textbf{0.00} & \textbf{0.00} & \textbf{0.00} \\
pulsar & \textbf{0.00} & \textbf{0.00} & \textbf{0.00} & \textbf{0.00} & \textbf{0.00} \\
catsvsdogs & \textbf{0.00} & \textbf{0.00} & \textbf{0.00} & \textbf{0.00} & \textbf{0.00} \\
bank & \textbf{\textcolor{gray}{0.01}} & \textbf{\textcolor{gray}{0.01}} & \textbf{\textcolor{gray}{0.01}} & \textbf{\textcolor{gray}{0.01}} & \textbf{\textcolor{gray}{0.01}} \\
default & \textbf{\textcolor{gray}{0.04}} & \textbf{\textcolor{gray}{0.04}} & \textbf{\textcolor{gray}{0.04}} & \textbf{\textcolor{gray}{0.04}} & \textbf{\textcolor{gray}{0.04}} \\
soft & \textbf{0.00} & \textbf{0.00} & \textbf{0.00} & \textbf{0.00} & \textbf{0.00} \\
adult & 0.08 & 0.08 & 0.08 & 0.08 & 0.08 \\
run & \textbf{\textcolor{gray}{0.03}} & \textbf{\textcolor{gray}{0.03}} & \textbf{\textcolor{gray}{0.02}} & \textbf{0.01} & \textbf{0.00} \\
smartgrid & \textbf{\textcolor{gray}{0.01}} & \textbf{0.01} & \textbf{0.00} & \textbf{0.00} & \textbf{0.00} \\
smoking & 0.10 & 0.10 & 0.10 & 0.10 & 0.10 \\
yearbook & \textbf{\textcolor{gray}{0.02}} & \textbf{0.01} & \textbf{0.01} & \textbf{0.01} & \textbf{0.00} \\
secondary & 0.33 & 0.33 & 0.33 & 0.33 & 0.33 \\
fashion & 0.42 & 0.34 & 0.30 & 0.24 & \textbf{\textcolor{gray}{0.05}} \\
cifar10 & 0.42 & 0.35 & 0.31 & 0.18 & \textbf{0.01} \\
mnist & 0.48 & 0.39 & 0.33 & 0.11 & \textbf{0.01} \\
letter & 0.70 & 0.70 & 0.70 & 0.70 & 0.70 \\
\bottomrule
\end{tabular}
\end{table}
\begin{table}[h]
\centering
\caption{Influence of non-i.i.d data with extreme drifts in $p(\x,y)$: test gaps between CRC using different sparse communication network topologies and RC at round $t=64$. 
}
\label{tab:topology_gap_XY}
\begin{tabular}{l|ccccc}
\toprule
\multicolumn{1}{c|}{} & \multicolumn{5}{c}{Topology of $\set{N}_l$} \\
Dataset & tree & +10 & +20 & +40 & +80 \\
\midrule
skin & \textbf{\textcolor{gray}{0.03}} & \textbf{\textcolor{gray}{0.02}} & \textbf{\textcolor{gray}{0.01}} & \textbf{0.01} & \textbf{0.00} \\
pulsar & \textbf{\textcolor{gray}{0.03}} & \textbf{\textcolor{gray}{0.02}} & \textbf{\textcolor{gray}{0.01}} & \textbf{0.00} & \textbf{0.00} \\
catsvsdogs & \textbf{0.01} & \textbf{0.00} & \textbf{0.00} & \textbf{0.00} & \textbf{0.00} \\
bank & \textbf{\textcolor{gray}{0.01}} & \textbf{\textcolor{gray}{0.01}} & \textbf{\textcolor{gray}{0.01}} & \textbf{\textcolor{gray}{0.01}} & \textbf{\textcolor{gray}{0.01}} \\
default & \textbf{\textcolor{gray}{0.04}} & \textbf{\textcolor{gray}{0.04}} & \textbf{\textcolor{gray}{0.04}} & \textbf{\textcolor{gray}{0.04}} & \textbf{\textcolor{gray}{0.04}} \\
soft & \textbf{0.00} & \textbf{0.00} & \textbf{0.00} & \textbf{0.00} & \textbf{0.00} \\
adult & 0.08 & 0.08 & 0.08 & 0.08 & 0.08 \\
run & 0.06 & \textbf{\textcolor{gray}{0.05}} & \textbf{\textcolor{gray}{0.03}} & \textbf{\textcolor{gray}{0.02}} & \textbf{0.01} \\
smartgrid & \textbf{\textcolor{gray}{0.02}} & \textbf{0.01} & \textbf{0.01} & \textbf{0.00} & \textbf{0.00} \\
smoking & 0.10 & 0.10 & 0.10 & 0.10 & 0.10 \\
yearbook & \textbf{\textcolor{gray}{0.04}} & \textbf{\textcolor{gray}{0.02}} & \textbf{\textcolor{gray}{0.02}} & \textbf{\textcolor{gray}{0.01}} & \textbf{0.01} \\
secondary & 0.33 & 0.33 & 0.33 & 0.33 & 0.33 \\
fashion & 0.44 & 0.36 & 0.30 & 0.25 & \textbf{\textcolor{gray}{0.02}} \\
cifar10 & 0.42 & 0.36 & 0.27 & 0.20 & \textbf{\textcolor{gray}{0.04}} \\
mnist & 0.53 & 0.45 & 0.38 & 0.14 & \textbf{0.01} \\
letter & 0.70 & 0.70 & 0.70 & 0.70 & 0.70 \\
\bottomrule
\end{tabular}
\end{table}

In despite of the extreme drifts, CRC is resilient and using sparse communication networks with just $11\%$ (+80) of possible edges is able to get gaps smaller than $0.01$ ($0.05$) in 14 (15), 9 (11) and 8 (12) out of 16 datasets with drifts in $p(\x)$, $p(y)$ and $p(\x,y)$, respectively. Notably, the effect of the extreme non-i.i.d. generated local training set can be drastically accommodated by adding a small fraction of the total edges.

\subsection{Size of local data, $m_v$}\label{sec:exper.m_v}

The availability of local data at each node directly impacts model quality. Larger datasets should intuitively lead to better estimates and faster convergence of CRC.  Table~\ref{tab:effect_train_size_global_gap} shows the effect of the training set size in the test error of RC.

In these experiments, we vary the size of local datasets $m_v \in \{20, 40, 80, 160, 320\}$ , while keeping the rest of default parameters. Table~\ref{tab:ml_gap_effect} shows the test gap between CRC for the different values of $m_v$ and their associated gold standards given by RC with $m= n \cdot m_v$ and $n=50$. The gold standard improves with $m_v$, since the amount of global training data $m$ increases linearly with the size of local training data $m_v$, and thus the test gaps between CRC and RC could increase with $m_v$ (see Table~\ref{tab:effect_train_size_global_gap}).

As the size of local training data $m_v$ increases, the gap between CRC and RC with $m= n \cdot m_v$ decreases. Even with only $m_v=20$ CRC is able to obtain gaps smaller than $0.01$ ($0.05)$ in $8$ (12) out of 16 with respect to RC using $m=1000$ samples, while for $m_v=320$, CRC obtain gaps smaller in 12 (15) datasets concerning RC using $16000$ samples.

\begin{table}[h]
\centering
\caption{Influence of local data size: test error gaps at round $t=64$ between CRC with different local training sizes $m_v$ and RC trained with $m=n\cdot m_v$ samples. 
}
\label{tab:ml_gap_effect}
\begin{tabular}{l|ccccc}
\toprule
\multicolumn{1}{c|}{} & \multicolumn{1}{c}{$m_v$=20} & \multicolumn{1}{c}{$m_v$=40} & \multicolumn{1}{c}{$m_v$=80} & \multicolumn{1}{c}{$m_v$=160} & \multicolumn{1}{c}{$m_v$=320} \\
Dataset & \multicolumn{1}{c}{$m$=1000} & \multicolumn{1}{c}{$m$=2000} & \multicolumn{1}{c}{$m$=4000} & \multicolumn{1}{c}{$m$=8000} & \multicolumn{1}{c}{$m$=16000} \\
\midrule
skin & \textbf{0.00} & \textbf{0.00} & \textbf{0.00} & \textbf{0.00} & \textbf{0.00} \\
pulsar & \textbf{0.00} & \textbf{0.00} & \textbf{0.00} & \textbf{0.00} & \textbf{0.00} \\
catsvsdogs & \textbf{0.00} & \textbf{0.00} & \textbf{0.00} & \textbf{0.00} & \textbf{0.00} \\
bank & \textbf{0.00} & \textbf{0.00} & \textbf{0.00} & \textbf{0.00} & \textbf{0.00} \\
default & \textbf{\textcolor{gray}{0.01}} & \textbf{0.01} & \textbf{0.01} & \textbf{0.00} & \textbf{0.00} \\
soft & \textbf{0.01} & \textbf{0.00} & \textbf{0.00} & \textbf{0.00} & \textbf{0.00} \\
adult & \textbf{\textcolor{gray}{0.02}} & \textbf{\textcolor{gray}{0.01}} & \textbf{0.01} & \textbf{0.00} & \textbf{0.00} \\
run & \textbf{0.01} & \textbf{0.01} & \textbf{0.00} & \textbf{0.00} & \textbf{0.00} \\
smartgrid & \textbf{0.01} & \textbf{0.00} & \textbf{0.00} & \textbf{0.00} & \textbf{0.00} \\
smoking & \textbf{\textcolor{gray}{0.02}} & \textbf{\textcolor{gray}{0.02}} & \textbf{\textcolor{gray}{0.01}} & \textbf{0.01} & \textbf{0.01} \\
yearbook & \textbf{0.01} & \textbf{\textcolor{gray}{0.01}} & \textbf{\textcolor{gray}{0.01}} & \textbf{\textcolor{gray}{0.01}} & \textbf{0.01} \\
secondary & \textbf{\textcolor{gray}{0.05}} & \textbf{\textcolor{gray}{0.02}} & \textbf{\textcolor{gray}{0.02}} & \textbf{\textcolor{gray}{0.01}} & \textbf{0.01} \\
fashion & 0.16 & 0.09 & \textbf{\textcolor{gray}{0.02}} & \textbf{\textcolor{gray}{0.02}} & \textbf{\textcolor{gray}{0.01}} \\
cifar10 & 0.28 & \textbf{\textcolor{gray}{0.01}} & \textbf{\textcolor{gray}{0.02}} & \textbf{\textcolor{gray}{0.02}} & \textbf{\textcolor{gray}{0.02}} \\
mnist & 0.66 & \textbf{\textcolor{gray}{0.03}} & 0.10 & \textbf{\textcolor{gray}{0.02}} & 0.10 \\
letter & 0.61 & 0.68 & 0.09 & 0.07 & \textbf{\textcolor{gray}{0.04}} \\
\bottomrule
\end{tabular}
\end{table}

\subsection{Size of communication network, $n$}\label{sec:exp.n}

Scalability is a key requirement for federated learning. It is therefore important to verify that CRC maintains performance as the number of users increases.

In these experiments we vary the size of the communication network $n \in \{20, 40, 80, 160, 320\}$ nodes, while keeping the rest of default parameters. The experiments drastically increase the difficulty of the learning scenarios in terms of the gaps between CRC and RC as $n$ increases due to the following reasons. The sparseness of a tree increases with $n$. For instance, for $n=20$ the edges of a tree are $n-1=19$ and they represent the $10\%$ of the total edges, while for $n=320$ they are less than $1\%$. As \( n \) increases, the average fraction of nodes within a given distance from any node decreases, making convergence more difficult
(see Figure~\ref{fig:distance.heatmap}). The global training set used by RC linearly increases with $n$ while the size of the local training sets remain constant $m_v=50$, which for $n=20$ has $m=1000$ samples and for $n=320$ has $m=16000$.

Table~\ref{tab:n_effect_gap} shows the test gaps between CRC and RC for different tree sizes, $n$. The test gap slightly increases as the size of the communication network $n$ increases: with the smallest communication network, $n=20$, the test gaps are smaller than $0.01$ ($0.05$) in 13 (15) out of 16, and with the largest $n=320$ in 7 (12) out of 16 datasets. These results highlight the robustness of CRC to large communication networks with hundreds of nodes.

\begin{table}[h]
\centering
\caption{Influence of communication network size: test error gaps between CRC and RC (best round per seed) for different network sizes $n$. 
}
\label{tab:n_effect_gap}
\begin{tabular}{l|ccccc}
\toprule
\multicolumn{1}{c|}{} & \multicolumn{1}{c}{$n$=20} & \multicolumn{1}{c}{$n$=40} & \multicolumn{1}{c}{$n$=80} & \multicolumn{1}{c}{$n$=160} & \multicolumn{1}{c}{$n$=320} \\
Dataset & $m$=1000 & $m$=2000 & $m$=4000 & $m$=8000 & $m$=16000 \\
\midrule
skin & \textbf{0.00} & \textbf{0.00} & \textbf{0.00} & \textbf{0.00} & \textbf{0.00} \\
pulsar & \textbf{0.00} & \textbf{0.00} & \textbf{0.00} & \textbf{0.00} & \textbf{0.00} \\
catsvsdogs & \textbf{0.00} & \textbf{0.00} & \textbf{0.00} & \textbf{0.00} & \textbf{0.00} \\
bank & \textbf{0.00} & \textbf{0.00} & \textbf{0.00} & \textbf{0.00} & \textbf{0.00} \\
default & \textbf{0.00} & \textbf{0.01} & \textbf{0.01} & \textbf{0.01} & \textbf{\textcolor{gray}{0.01}} \\
soft & \textbf{0.00} & \textbf{0.00} & \textbf{0.00} & \textbf{0.00} & \textbf{0.00} \\
adult & \textbf{0.01} & \textbf{0.01} & \textbf{0.01} & \textbf{\textcolor{gray}{0.01}} & \textbf{\textcolor{gray}{0.01}} \\
run & \textbf{0.00} & \textbf{0.01} & \textbf{0.00} & \textbf{0.00} & \textbf{0.00} \\
smartgrid & \textbf{0.00} & \textbf{0.00} & \textbf{0.00} & \textbf{0.00} & \textbf{0.00} \\
smoking & \textbf{\textcolor{gray}{0.01}} & \textbf{\textcolor{gray}{0.02}} & \textbf{\textcolor{gray}{0.02}} & \textbf{\textcolor{gray}{0.02}} & \textbf{\textcolor{gray}{0.02}} \\
yearbook & \textbf{0.00} & \textbf{0.01} & \textbf{\textcolor{gray}{0.02}} & \textbf{\textcolor{gray}{0.02}} & \textbf{\textcolor{gray}{0.03}} \\
secondary & \textbf{\textcolor{gray}{0.02}} & \textbf{\textcolor{gray}{0.02}} & \textbf{\textcolor{gray}{0.03}} & \textbf{\textcolor{gray}{0.03}} & \textbf{\textcolor{gray}{0.03}} \\
fashion & \textbf{0.01} & \textbf{\textcolor{gray}{0.01}} & \textbf{\textcolor{gray}{0.03}} & 0.07 & 0.06 \\
cifar10 & \textbf{0.01} & \textbf{\textcolor{gray}{0.01}} & \textbf{\textcolor{gray}{0.02}} & \textbf{\textcolor{gray}{0.04}} & 0.10 \\
mnist & \textbf{0.00} & \textbf{\textcolor{gray}{0.02}} & 0.06 & 0.08 & 0.14 \\
letter & 0.39 & 0.50 & 0.28 & 0.28 & 0.24 \\
\bottomrule
\end{tabular}
\end{table}

\subsection{Fragmentation of global training data, $n \cdot m_v$ constant}

In this section, we analyze the impact of splitting a fixed global training set across communication networks of varying sizes. This experiment is particularly valuable for understanding how data fragmentation influences learning in decentralized networks. 

Given a fixed training data of size $m$, and a federated decentralized scenario with $n$ users, the data fragmentation increases with $n$ since the corresponding $m_v= m/n$. Note that the fragmentation does not affect RC because the global training data remains the same with independence of the value of $n$, and thus the gold-standard is the same. In these experiments, we vary the number of users $n=20, 40, 80, 160, 320$ while the training set decreases linearly so as to the global training set remains constant. All other parameters are set to their default values.As \textit{n} increases, the experiments become more challenging due to greater tree sparsity and a reduced average fraction of nodes within a given distance from any node (see Section 4.2). The difficulty is further amplified by the linear decrease in the size of local training sets, while the global training set remains fixed at \textit{m} = 12800.

Table~\ref{tab:fragmentation_gap_effect} shows the test gaps between CRC for different values of $n$, using local training samples of size $m/n$, and RC with a global training set of $m = 12800$ samples. In the most favorable conditions ($n=20$, $m_v=640$), the CRC achieves test gaps smaller than $0.01$ ($0.05$) in 15 (16) out of 16, while in the most extreme case of data fragmentation ($n = 320$, $m_v = 40$), the algorithm is still able to achieve test gaps below $0.01$ ($0.05$) in 7 (12) out of 16 datasets.

\begin{table}[h]
\centering
\caption{Influence of global data fragmentation: test error gaps of CRC with respect to RC, trained with $m$=12800, for different levels of data fragmentation. 
}
\label{tab:fragmentation_gap_effect}
\begin{tabular}{l|ccccc}
\toprule
\multicolumn{1}{c|}{} & \multicolumn{1}{c}{$n$=20} & \multicolumn{1}{c}{$n$=40} & \multicolumn{1}{c}{$n$=80} & \multicolumn{1}{c}{$n$=160} & \multicolumn{1}{c}{$n$=320} \\
Dataset & $m_l$=640 & $m_l$=320 & $m_l$=160 & $m_l$=80 & $m_l$=40 \\
\midrule
skin & \textbf{0.00} & \textbf{0.00} & \textbf{0.00} & \textbf{0.00} & \textbf{0.00} \\
pulsar & \textbf{0.00} & \textbf{0.00} & \textbf{0.00} & \textbf{0.00} & \textbf{0.00} \\
catsvsdogs & \textbf{0.00} & \textbf{0.00} & \textbf{0.00} & \textbf{0.00} & \textbf{0.00} \\
bank & \textbf{0.00} & \textbf{0.00} & \textbf{0.00} & \textbf{0.00} & \textbf{0.00} \\
default & \textbf{0.00} & \textbf{0.00} & \textbf{0.00} & \textbf{0.01} & \textbf{\textcolor{gray}{0.01}} \\
soft & \textbf{0.00} & \textbf{0.00} & \textbf{0.00} & \textbf{0.00} & \textbf{0.00} \\
adult & \textbf{0.00} & \textbf{0.00} & \textbf{0.00} & \textbf{0.01} & \textbf{\textcolor{gray}{0.01}} \\
run & \textbf{0.00} & \textbf{0.00} & \textbf{0.00} & \textbf{0.00} & \textbf{0.00} \\
smartgrid & \textbf{0.00} & \textbf{0.00} & \textbf{0.00} & \textbf{0.00} & \textbf{0.00} \\
smoking & \textbf{0.00} & \textbf{0.01} & \textbf{0.01} & \textbf{\textcolor{gray}{0.02}} & \textbf{\textcolor{gray}{0.03}} \\
yearbook & \textbf{0.00} & \textbf{0.01} & \textbf{\textcolor{gray}{0.01}} & \textbf{\textcolor{gray}{0.02}} & \textbf{\textcolor{gray}{0.03}} \\
secondary & \textbf{0.00} & \textbf{0.01} & \textbf{\textcolor{gray}{0.01}} & \textbf{\textcolor{gray}{0.02}} & \textbf{\textcolor{gray}{0.04}} \\
fashion & \textbf{0.00} & \textbf{0.01} & \textbf{\textcolor{gray}{0.02}} & \textbf{\textcolor{gray}{0.03}} & 0.11 \\
cifar10 & \textbf{0.01} & \textbf{\textcolor{gray}{0.02}} & \textbf{\textcolor{gray}{0.03}} & \textbf{\textcolor{gray}{0.04}} & 0.09 \\
mnist & \textbf{0.01} & \textbf{\textcolor{gray}{0.01}} & \textbf{\textcolor{gray}{0.02}} & 0.07 & 0.17 \\
letter & \textbf{\textcolor{gray}{0.02}} & \textbf{\textcolor{gray}{0.04}} & 0.07 & 0.11 & 0.74 \\
\bottomrule
\end{tabular}
\end{table}

\subsection{Number of local iterations in each round, $iter$}

Performing multiple local updates before communicating could accelerate convergence by reducing the frequency of costly communication rounds. However, this strategy may also introduce instability in the learning process, as local models can diverge from each other during independent updates. Thus, in this section, we analyze the impact of varying the number of local iterations $iter \in \{1,2,3\}$ of CRC, where the rest are the default parameters.

Table~\ref{tab:iterations_gap_effect} shows the test gaps between CRC for different number of iterations, $iter$, and the RC. In particular, it shows for each value of $iter$ the round $t$ in which the gap is lower than $0.01$ and in case that CRC is not able to reach a gap lower than $0.01$ the gap at round 64. The results show that increasing the number of local iterations ($iter$) can speed up the convergence, reducing the number of communication rounds to achieve a gap $< 0.01$. 

\begin{table}[h]
\centering
\caption{Influence of number of local iterations: test error gaps, and round $t$ where gap $<$ 0.01 (if any), for different number of local iterations. 
}
\label{tab:iterations_gap_effect}
\begin{tabular}{l|cr|cr|cr}
\toprule
 & \multicolumn{2}{c|}{$iter$=1} & \multicolumn{2}{c|}{$iter$=2} & \multicolumn{2}{c}{$iter$=3} \\
Dataset & gap & $t$ & gap & $t$ & gap & $t$ \\
\midrule
skin & \textbf{0.01} & \textbf{32} & \textbf{0.01} & \textbf{16} & \textbf{0.01} & \textbf{11} \\
pulsar & \textbf{0.01} & \textbf{13} & \textbf{0.01} & \textbf{8} & \textbf{0.01} & \textbf{7} \\
catsvsdogs & \textbf{0.01} & \textbf{5} & \textbf{0.01} & \textbf{4} & \textbf{0.01} & \textbf{3} \\
bank & \textbf{0.01} & \textbf{11} & \textbf{0.01} & \textbf{6} & \textbf{0.01} & \textbf{5} \\
default & \textbf{0.01} & \textbf{29} & \textbf{\textcolor{gray}{0.01}} & \textbf{\textcolor{gray}{64}} & \textbf{\textcolor{gray}{0.02}} & \textbf{\textcolor{gray}{64}} \\
soft & \textbf{0.00} & \textbf{2} & \textbf{0.00} & \textbf{1} & \textbf{0.00} & \textbf{1} \\
adult & \textbf{0.01} & \textbf{56} & \textbf{0.01} & \textbf{43} & \textbf{0.01} & \textbf{45} \\
run & \textbf{0.01} & \textbf{48} & \textbf{0.01} & \textbf{25} & \textbf{0.01} & \textbf{18} \\
smartgrid & \textbf{0.01} & \textbf{42} & \textbf{0.01} & \textbf{24} & \textbf{0.01} & \textbf{33} \\
smoking & \textbf{\textcolor{gray}{0.02}} & \textbf{\textcolor{gray}{64}} & \textbf{\textcolor{gray}{0.01}} & \textbf{\textcolor{gray}{64}} & \textbf{\textcolor{gray}{0.02}} & \textbf{\textcolor{gray}{64}} \\
yearbook & \textbf{\textcolor{gray}{0.01}} & \textbf{\textcolor{gray}{64}} & \textbf{\textcolor{gray}{0.01}} & \textbf{\textcolor{gray}{64}} & \textbf{\textcolor{gray}{0.01}} & \textbf{\textcolor{gray}{64}} \\
secondary & \textbf{\textcolor{gray}{0.03}} & \textbf{\textcolor{gray}{64}} & \textbf{\textcolor{gray}{0.02}} & \textbf{\textcolor{gray}{64}} & \textbf{\textcolor{gray}{0.01}} & \textbf{\textcolor{gray}{64}} \\
fashion & \textbf{\textcolor{gray}{0.02}} & \textbf{\textcolor{gray}{64}} & 0.09 & 64 & \textbf{\textcolor{gray}{0.02}} & \textbf{\textcolor{gray}{64}} \\
cifar10 & \textbf{\textcolor{gray}{0.02}} & \textbf{\textcolor{gray}{64}} & \textbf{\textcolor{gray}{0.02}} & \textbf{\textcolor{gray}{64}} & \textbf{\textcolor{gray}{0.02}} & \textbf{\textcolor{gray}{64}} \\
mnist & 0.10 & 64 & 0.26 & 64 & 0.34 & 64 \\
letter & 0.59 & 64 & 0.59 & 64 & 0.58 & 64 \\
\bottomrule
\end{tabular}
\end{table}

\subsection{Dynamic changes in the communication network}

Finally, we study the impact of dynamic changes in the communication network, which are common in many real-world scenarios due to user mobility or network variability. In the experiments, we analyze the effect of a communication network that changes with different periods $\delta$, which determines how frequently the network edges are changed, where the rest are the default parameters.

Table~\ref{tab:period_gap_effect} shows the test gaps between CRC and RC. In these results, for $\delta \in \{1,2,4,8\}$, the set of connections between nodes is updated every $\delta$ communication rounds, while for $\delta=\infty$, the network remains static. In 9 out of 16 datasets, CRC achieves gaps smaller than $0.01$ with both dynamic and static communication networks. In 6 out of 16 dynamic communication networks can reduce the gap compared to the static case, while in only 1 dataset the gap increases under dynamic conditions. These results suggest that CRC can benefit from dynamic communication networks, because dynamic networks can reduce effective communication distances among nodes, reinforcing CRC’s robustness.

\begin{table}[h]
\centering
\caption{Influence of dynamic changes in the communication network: test error gaps between CRC and RC at round $t=64$ as a function of the local update period $\delta$.}
\label{tab:period_gap_effect}
\begin{tabular}{l|ccccc}
\toprule
Dataset & $\delta$=$\infty$ & $\delta$=8 & $\delta$=4 & $\delta$=2 & $\delta$=1 \\
\midrule
skin & \textbf{0.00} & \textbf{0.00} & \textbf{0.00} & \textbf{0.00} & \textbf{0.00} \\
pulsar & \textbf{0.00} & \textbf{0.00} & \textbf{0.00} & \textbf{0.00} & \textbf{0.00} \\
catsvsdogs & \textbf{0.00} & \textbf{0.00} & \textbf{0.00} & \textbf{0.00} & \textbf{0.00} \\
bank & \textbf{0.00} & \textbf{0.00} & \textbf{0.00} & \textbf{0.00} & \textbf{0.00} \\
default & \textbf{0.01} & \textbf{0.01} & \textbf{0.00} & \textbf{0.01} & \textbf{0.01} \\
soft & \textbf{0.00} & \textbf{0.00} & \textbf{0.00} & \textbf{0.00} & \textbf{0.00} \\
adult & \textbf{0.01} & \textbf{0.01} & \textbf{0.00} & \textbf{0.00} & \textbf{0.00} \\
run & \textbf{0.00} & \textbf{0.00} & \textbf{0.00} & \textbf{0.00} & \textbf{0.00} \\
smartgrid & \textbf{0.00} & \textbf{0.00} & \textbf{0.00} & \textbf{0.00} & \textbf{0.00} \\
smoking & \textcolor{gray}{\textbf{0.02}} & \textcolor{gray}{\textbf{0.01}} & \textbf{0.01} & \textbf{0.01} & \textbf{0.01} \\
yearbook & \textcolor{gray}{\textbf{0.01}} & \textbf{0.00} & \textbf{0.00} & \textbf{0.01} & \textbf{0.01} \\
secondary & \textcolor{gray}{\textbf{0.03}} & \textcolor{gray}{\textbf{0.02}} & \textcolor{gray}{\textbf{0.02}} & \textbf{0.01} & \textcolor{gray}{\textbf{0.01}} \\
fashion & \textcolor{gray}{\textbf{0.02}} & \textcolor{gray}{\textbf{0.04}} & \textbf{0.01} & 0.44 & 0.45 \\
cifar10 & \textcolor{gray}{\textbf{0.02}} & \textbf{0.00} & \textbf{0.00} & 0.14 & 0.43 \\
mnist & 0.10 & 0.57 & 0.34 & 0.34 & 0.67 \\
letter & 0.59 & 0.57 & 0.57 & 0.57 & 0.57 \\
\bottomrule
\end{tabular}
\end{table}

\section{Conclusions}
\label{sec:conclusions}

The current research proposes a decentralized federated learning implementation of risk-based calibration (RC) algorithm, called collaborative risk-based calibration (CRC). The goal of CRC is to minimize the empirical error of probabilistic generative classifiers over arbitrary communication network of data owners. CRC learns local models by sharing statistics between neighbors that are iteratively updated using local training datasets in a decentralized way. CRC can be applied in any network setting defined over an undirected graph that can change over time, and to any probabilistic generative classifier with a closed-form learning algorithm of the model parameters from statistics.

Despite the simplicity of CRC, we prove that it is able to replicate the RC algorithm with all the training data in the network using a fully connected communication network. Through extensive experimentation in 16 datasets, we    analyze the behavior of CRC in trees, the sparsest communication networks with a single connected components, and measure the error difference (gap) with respect to the gold standard represented by RC using all the available training data in the network. The experiments are designed to imposed challenging conditions as a baseline, including limited data per node and sparsely connected networks; when evaluating non-i.i.d. scenarios, we set extreme conditions to clearly assess the robustness of CRC. The main findings can be summarized as follows: 
\begin{itemize}
    \item \textbf{Convergence of local models:} CRC achieves convergence to the global RC model with remarkable accuracy and consistency, outperforming ML in nearly all datasets while maintaining extremely low variance across local models.

    \item \textbf{Sparse communication networks:} CRC demonstrates strong resilience to sparse topologies, achieving high accuracy even with minimal connectivity, and further improving with only a small increase in edges.

    \item \textbf{Non-i.i.d. data:} CRC remains robust under severe non-i.i.d. conditions, recovering performance close to centralized RC in most datasets when just a small fraction of additional edges is added to the network.

    \item \textbf{Size of local data:} the gap between CRC and RC decreases as the local training data increases.

    \item \textbf{Size of communication network:} CRC maintains competitive performance across increasing network sizes, showcasing its scalability and robustness even under highly sparse and challenging network conditions.

    \item \textbf{Fragmentation of global training data:} CRC effectively handles extreme data fragmentation, preserving low error gaps with RC even when the global dataset is split across hundreds of nodes.

    \item \textbf{Number of local iterations per round:} Increasing local iterations can accelerate CRC’s convergence, enabling faster approach to the RC reference with fewer communication rounds.

    \item \textbf{Dynamic changes in the communication network:} CRC benefits from dynamic communication networks, often improving accuracy by reducing communication distances and enhancing resilience to topology variability.
\end{itemize}

All the classifiers, learning algorithms, datasets and experiments included in this paper can be found in the open-source library developed in Python \url{https://gitlab.bcamath.org/aperez/decentralized_risk-based_calibration}.

\appendices

\section{The naive Bayes classifier}\label{app:nb}
A popular example of probabilistic generative classifier based on Bayesian networks is naive Bayes (NB). In this work, we consider NB based on conditional Gaussian Bayesian networks which allows to deal with both continuous and discrete features distribution. In particular, NB based on the following factorization of the joint probability distribution:
\begin{equation}
p_{\tht}(\x,y)= p_{\tht}(y)\prod_{i=1}^d p_{\tht}(x_i|y),\label{eq:nb}
\end{equation} 
where $p_{\tht}(x_i|y)$ follows a Gaussian distribution or a categorical distribution for each $y$ when $x_i$ is a continuous or discrete feature, respectively. NB assumes that the features are conditionally independent given the class, which effectively controls the number of parameters of the probabilistic model and thus the overfitting. Despite of the strong independence assumption of NB, it has shown good performance in many practical scenarios~\cite{domingos1997optimalityNB, rish2001empiricalNB, zhang2004optimalityNB}. Moreover, in \cite{greiner02} the authors show that for continuous variables NB and logistic regression are the same class of models when we impose additional constraints on the variance of the Gaussian distributions $p_{\tht}(x_i|y)$ (homocedasticity assumption).

Maximum likelihood learning algorithm of NB is closed form. The statistics mapping can be decomposed into components corresponding to factor from \ref{eq:nb} including $p_{\tht}(y)$ and each the class conditional distribution of each feature $p_{\tht}(x_i|y)$ for $i=1,...,d$, $s(\x,y)=(s_0(y),s_1(x_1,y),...,s_d(x_d,y))$. The statistics for the class $s_0(y)=\bd{e}_y$ is the $y$-th vector in the standard basis of $\Real^{r}$, and $s_i$ depends on the continuous or discrete nature of the feature $x_i$.

When the $i$-th input feature $x_i$ is discrete, we denote its support as $\set{X}_i=\{1,...,r_i\}$ and $p_{\tht}(x_i|y)$ is a categorical distribution for $y \in \set{Y}$. The component $s_i(x_i,y)=\bd{e}_y \otimes \bd{e}_{x_i}$ where $\otimes$ is the Kronecker product, and it represents the counting statistics associated to each value of $(x_i,y) \in (\set{X}_i,\set{Y})$. When the $i$-th feature is continuous, its support is $\Real$ and $p_{\tht}(x_i|y)$ is an univariate Gaussian distribution with mean $\mu_y$ and variance $\sigma_y^2$ for $y \in \set{Y}$. The component $s_i(x_i,y)=\bd{e}_y \otimes (1,x_i,x_i^2)$ allows the computation of the zeroth, first, and second order moments required to get the maximum likelihood parameters of the Gaussian distribution for each value of $y$. 

Then the mapping $\theta(\bd{s})$ is given by $p_{\tht}(y)=s_{0,y}/\sum_{y' \in \set{Y}}s_{0,y'}$ for the probabilities of the categorical distribution of the class; by  $p_{\tht}(x_i|y)=s_{i,y,x_i}/\sum_{x_i' \in \set{X}_i}s_{i,y,x_i'}$ for the probabilities of the categorical distribution of $x_i$ given the class $y \in \set{Y}$; and by $\mu_{i,y}= s_{i,y,2}/s_{i,y,1}$ and $\sigma_{i,y}^2= s_{i,y,3}/s_{i,y,1} - \mu_{i,y} \cdot \mu_{i,y}^T$, for the mean and variance of the Gaussian density function of $x_i$ given the class label $y \in \set{Y}$.

\subsection{Uniform initialization}
On the experiments, we use the same initialization of the local statistics in all the nodes: the counting statistics for the class are $m_y=m^0/r$ for $y=1,...,r$, the counting statistics for discrete variables are $m^0/(r \cdot r_i)$ for $x_i=1,...,r_i$ and $y=1,...,r$, the statistics for the zeroth, first and second moments of continuous variables are $m^0/r$, $m^0/r$ and $0$, respectively. This leads to a classifier with uniform probabilities, $p_{\tht}(y|\x)$ for all $\x \in \X$, with $\tht= \theta(\bd{s})$.

\section{CRC with full communication network}

This section, proves the equivalence between CRC with full communication network and RC without requiring further assumptions on the local data generation process.

\repeattheorem{thm:equivalence}{
Let $(\set{V},\set{E})$ be a communication network with $\set{N}_v = \set{V}$ for all $v \in \set{V}$. Let $\{(X_v,Y_v): v \in \set{V}\}$ be a partition of the global training data $(X,Y)$, where $m = \sum_{v\in \set{V}} m_v$. Consider RC with learning rate $lr > 0$, using the global training data $(X,Y)$ with uniform initialization and a single iteration, i.e., $iter = 1$. Now consider CRC with equivalent sample size $m^0 = \frac{m}{lr \cdot n}$, using the local data $(X_v,Y_v)$ for each $v \in \set{V}$ with uniform initialization and one iteration ($iter = 1$). Then, all local parameters in the aggregation step of CRC coincide with the parameters of RC at the previous iteration: 
$$\theta(\bar{\bd{s}}^{t}_v) = \theta(\bd{s}^{t-1}_g), \quad \text{for } t = 1, \dots, t^{\max}, \text{ and } v \in \set{V}.$$}

\begin{proof}
We proceed by induction on the local aggregated statistic $\bar{\bd{s}}_v^{t}$ for $v \in \set{V}$.

We first observe the following:
\begin{enumerate}
    \item All aggregated statistics are identical across nodes: $\bar{\bd{s}}^{t}_v = \bar{\bd{s}}^{t}$ for all $v \in \set{V}$.
    \item The aggregated statistics $\bar{\bd{s}}^t$ are updated locally using $(X_v, Y_v)$ and a classifier with parameters $\theta(\bar{\bd{s}}^{t})$.
    \item The parameters of a classifier obtained from a statistic $\bd{s}$ are invariant under positive rescaling of $\bd{s}$.
\end{enumerate}

\textbf{Base case ($t = 1$):}  
At initialization, all local statistics are equal to $\bd{s}^0$, and the global statistic $\bd{s}_g^0$ is a rescaled version of the local one:
$$
\bd{s}^{0} = \text{initialize}(m^0), \quad \bd{s}^{0}_g = \text{initialize}(m) = \frac{m}{m^0} \cdot \bd{s}^{0}.
$$
Then, at round $t = 1$, the aggregated local statistics are $\bar{\bd{s}}^1 = \bd{s}^0$ for all $v \in \set{V}$, and thus $\theta(\bd{s}_g^0) = \theta(\bar{\bd{s}}^1)$.

\textbf{Inductive step:}  
Assume that at round $t$, the aggregated local statistics are $\bar{\bd{s}}^{t}$ for all $v \in \set{V}$, and the global statistic at $t-1$ satisfies $\bd{s}_g^{t-1} = \frac{m}{m^0} \cdot \bar{\bd{s}}^t$. Then both RC and CRC compute the same classifier parameters, $\theta(\bar{\bd{s}}^{t}) = \theta(\bd{s}_g^{t-1}) = \tht$.

Next, compute the RC and CRC updates with parameter $\tht$ for each node $v \in \set{V}$:
\begin{align}
\bd{s}_v^{t} &= \bar{\bd{s}}^{t} + s(X_v, Y_v) - s(X_v, \tht),\\
\bd{s}_g^{t} &= \bd{s}_g^{t-1} + lr \cdot \left( s(X, Y) - s(X, \tht) \right).
\end{align}

At the beginning of round $t+1$, compute the CRC aggregation step for each node $v$:
\begin{align}
\bar{\bd{s}}_v^{t+1} &= \frac{1}{n} \sum_{v \in \set{V}} \bd{s}_v^t\\
&= \bar{\bd{s}}^{t} + \frac{1}{n} \sum_{v \in \set{V}} \left( s(X_v, Y_v) - s(X_v, \tht) \right)\\
&= \bar{\bd{s}}^{t} + \frac{1}{n} \left( s(X, Y) - s(X, \tht) \right),
\end{align}
where the last equality holds because the statistic mapping is additive. This final expression does not depend on node $v$, so the aggregated statistics at round $t+1$ are the same for all $v \in \set{V}$, i.e., $\bar{\bd{s}}^{t+1}$.

Multiplying both sides by $\frac{m}{m^0}$, we obtain:
$$
\frac{m}{m^0} \cdot \bar{\bd{s}}^{t+1} = \bd{s}_g^{t-1} + \frac{m}{m^0 \cdot n} \left( s(X,Y) - s(X,\tht) \right) = \bd{s}_g^t,
$$
for $lr = \frac{m}{m^0 \cdot n}$.
\end{proof}

This result proves that by selecting the equivalent sample size $m^0 = \frac{m}{lr} \cdot n$, CRC with full communication network is equivalent to RC with a learning rate of $lr$ just by assuming that RC uses the global training data available at the network of users.

\bibliographystyle{IEEEtran}
\bibliography{biblioCRC.bib}

\begin{thebibliography}{10}
\providecommand{\url}[1]{#1}
\csname url@samestyle\endcsname
\providecommand{\newblock}{\relax}
\providecommand{\bibinfo}[2]{#2}
\providecommand{\BIBentrySTDinterwordspacing}{\spaceskip=0pt\relax}
\providecommand{\BIBentryALTinterwordstretchfactor}{4}
\providecommand{\BIBentryALTinterwordspacing}{\spaceskip=\fontdimen2\font plus
\BIBentryALTinterwordstretchfactor\fontdimen3\font minus \fontdimen4\font\relax}
\providecommand{\BIBforeignlanguage}[2]{{%
\expandafter\ifx\csname l@#1\endcsname\relax
\typeout{** WARNING: IEEEtran.bst: No hyphenation pattern has been}%
\typeout{** loaded for the language `#1'. Using the pattern for}%
\typeout{** the default language instead.}%
\else
\language=\csname l@#1\endcsname
\fi
#2}}
\providecommand{\BIBdecl}{\relax}
\BIBdecl

\bibitem{mcmahan17}
B.~McMahan, E.~Moore, D.~Ramage, S.~Hampson, and B.~A. y~Arcas, ``Communication-efficient learning of deep networks from decentralized data,'' in \emph{Proceedings of the 20th International Conference on Artificial Intelligence and Statistics}, vol.~54, 2017.

\bibitem{Yang19}
Q.~Yang, Y.~Liu, T.~Chen, and Y.~Tong, ``Federated machine learning: Concept and applications,'' \emph{ACM Transactions on Intelligent Systems and Technology}, vol.~10, no.~2, pp. 1--19, 2019.

\bibitem{Jakub16}
J.~Konečný, H.~B. McMahan, F.~X. Yu, P.~Richtarik, A.~T. Suresh, and D.~Bacon, ``Federated learning: Strategies for improving communication efficiency,'' in \emph{NIPS Workshop on Private Multi-Party Machine Learning}, 2016.

\bibitem{Sun23}
T.~Sun, D.~Li, and B.~Wang, ``Decentralized federated averaging,'' \emph{IEEE Transactions on Pattern Analysis and Machine Intelligence}, vol.~4, no.~5, pp. 4289 -- 4301, 2023.

\bibitem{mcmahan2020communication}
H.~B. McMahan, D.~M. Bacon, J.~Konecny, and X.~Yu, ``Communication efficient federated learning,'' may 2020.

\bibitem{Granqvist24}
F.~Granqvist, C.~Song, A.~Cahill, R.~van Dalen, M.~Pelikan, Y.~S. Chan, X.~Feng, N.~Krishnaswami, V.~Jina, and M.~Chitnis, ``pfl-research: simulation framework for accelerating research in private federated learning,'' in \emph{Advances in Neural Information Processing Systems}, vol.~37, 2024.

\bibitem{Li20}
T.~Li, A.~K. Sahu, A.~Talwalkar, and V.~Smith, ``Federated learning: Challenges, methods, and future directions,'' \emph{IEEE Signal Processing Magazine}, vol.~37, no.~3, pp. 50--60, 2020.

\bibitem{Kairouz21}
P.~Kairouz \emph{et~al.}, \emph{Advances and Open Problems in Federated Learning}.\hskip 1em plus 0.5em minus 0.4em\relax Now Foundations and Trends, 2021, vol.~14, no. 1--2.

\bibitem{lalitha2019decentralized}
A.~Lalitha, X.~Wang, O.~Kilinc, Y.~Lu, T.~Javidi, and F.~Koushanfar, ``Decentralized {B}ayesian learning over graphs,'' \emph{arXiv preprint}, 2019.

\bibitem{kalra2023}
S.~Kalra, J.~Wen, J.~C. Cresswell, M.~Volkovs, and H.~R. Tizhoosh, ``Decentralized federated learning through proxy model sharing,'' \emph{Nature Communications}, vol.~14, no. 2899, 2023.

\bibitem{BANABILAH2022}
S.~Banabilah, M.~Aloqaily, E.~Alsayed, N.~Malik, and Y.~Jararweh, ``Federated learning review: Fundamentals, enabling technologies, and future applications,'' \emph{Information Processing \& Management}, 2022.

\bibitem{GAMAZOREAL2023}
J.-C. Gamazo-Real, R.~{Torres Fernández}, and A.~{Murillo Armas}, ``Comparison of edge computing methods in internet of things architectures for efficient estimation of indoor environmental parameters with machine learning,'' \emph{Engineering Applications of Artificial Intelligence}, vol. 126, no. 107149, 2023.

\bibitem{Antunes22}
R.~S. Antunes, C.~Andr\'{e}~da Costa, A.~K\"{u}derle, I.~A. Yari, and B.~Eskofier, ``Federated learning for healthcare: Systematic review and architecture proposal,'' \emph{ACM Transactions on Intelligent Systems and Technology}, vol.~13, no.~4, 2022.

\bibitem{JOCHEMS2017}
A.~Jochems, T.~M. Deist, I.~{El Naqa}, M.~Kessler, C.~Mayo, J.~Reeves, S.~Jolly, M.~Matuszak, R.~{Ten Haken}, J.~{van Soest}, C.~Oberije, C.~Faivre-Finn, G.~Price, D.~{de Ruysscher}, P.~Lambin, and A.~Dekker, ``Developing and validating a survival prediction model for {NSCLC} patients through distributed learning across 3 countries,'' \emph{International Journal of Radiation Oncology*Biology*Physics}, vol.~99, no.~2, pp. 344--352, 2017.

\bibitem{Pezoulas2020}
V.~C. Pezoulas, K.~D. Kourou, F.~Kalatzis, T.~P. Exarchos, E.~Zampeli, S.~Gandolfo, A.~Goules, C.~Baldini, F.~Skopouli, S.~De~Vita, A.~G. Tzioufas, and D.~I. Fotiadis, ``Overcoming the barriers that obscure the interlinking and analysis of clinical data through harmonization and incremental learning,'' \emph{IEEE Open Journal of Engineering in Medicine and Biology}, vol.~1, pp. 83--90, 2020.

\bibitem{NIPS:Smith17}
V.~Smith, C.-K. Chiang, M.~Sanjabi, and A.~Talwalkar, ``Federated multi-task learning,'' in \emph{Proceedings of the 31st International Conference on Neural Information Processing Systems}, vol.~30, 2017.

\bibitem{Navia22}
A.~Navia-V\'{a}zquez, R.~D\'{\i}az-Morales, and M.~Fern\'{a}ndez-D\'{\i}az, ``Budget distributed support vector machine for non-id federated learning scenarios,'' \emph{ACM Trans. Intell. Syst. Technol.}, vol.~13, no.~6, 2022.

\bibitem{boroujeni2025}
M.~G. Boroujeni, A.~Krause, and G.~F. Trecate, ``Personalized federated learning of probabilistic models: A {PAC}-{B}ayesian approach,'' \emph{arXiv preprint}, 2025.

\bibitem{ashman2022}
M.~Ashman, T.~D. Bui, C.~V. Nguyen, S.~Markou, A.~Weller, S.~Swaroop, and R.~E. Turner, ``Partitioned variational inference: A framework for probabilistic federated learning,'' \emph{arXiv preprint}, 2022.

\bibitem{perez2025}
A.~Pérez, C.~Echegoyen, and G.~Santafé, ``Risk-based calibration for generative classifiers,'' \emph{arXiv preprint}, 2025.

\bibitem{bielza2014discrete}
C.~Bielza and P.~Larra\~{n}aga, ``Discrete {B}ayesian network classifiers: A survey,'' \emph{ACM Computing Surveys}, vol.~47, no.~1, 2014.

\bibitem{perez2006gaussian}
A.~P{\'e}rez, P.~Larra{\~n}aga, and I.~Inza, ``Supervised classification with conditional {G}aussian networks: Increasing the structure complexity from naive {B}ayes,'' \emph{International Journal of Approximate Reasoning}, vol.~43, no.~1, pp. 1--25, 2006.

\bibitem{augenstein2019generative}
S.~Augenstein, H.~B. McMahan, D.~Ramage, S.~Ramaswamy, P.~Kairouz, M.~Chen, R.~Mathews, and B.~A. y~Arcas, ``Generative models for effective {ML} on private, decentralized datasets,'' in \emph{International Conference on Learning Representations}, 2020.

\bibitem{GM20}
H.~GM, M.~K. Gourisaria, M.~Pandey, and S.~S. Rautaray, ``A comprehensive survey and analysis of generative models in machine learning,'' \emph{Computer Science Review}, vol.~38, p. 100285, 2020.

\bibitem{ng2001discriminative}
A.~Ng and M.~Jordan, ``On discriminative vs. generative classifiers: A comparison of logistic regression and naive {B}ayes,'' in \emph{Advances in Neural Information Processing Systems}, vol.~14, 2001.

\bibitem{Cuesta2019}
Y.~Sun, A.~Cuesta-Infante, and K.~Veeramachaneni, ``Learning vine copula models for synthetic data generation,'' \emph{AAAI Conference on Artificial Intelligence}, vol.~33, no.~01, 2019.

\bibitem{dempster1977}
A.~P. Dempster, N.~M. Laird, and D.~B. Rubin, ``Maximum likelihood from incomplete data via the {EM} algorithm,'' \emph{Journal of the Royal Statistical Society: Series B}, vol.~39, no.~1, pp. 1--22, 1977.

\bibitem{Tolou23}
T.~Shadbahr, M.~Roberts, J.~Stanczuk, J.~Gilbey, P.~Teare, S.~Dittmer, M.~Thorpe, R.~Torn{\'e}, E.~Sala, P.~Li{\'o}, M.~Patel, J.~Preller, I.~Selby, A.~Breger, J.~Weir-McCall, E.~Gkrania-Klotsas, A.~Korhonen, E.~Jefferson, G.~Langs, G.~Yang, H.~Prosch, J.~Babar, L.~{Escudero S{\'a}nchez}, M.~Wassin, M.~Holzer, N.~Walton, J.~Rudd, T.~Mirtti, A.~Rannikko, J.~Aston, J.~Tang, and C.-B. Sch{\"o}nlieb, ``The impact of imputation quality on machine learning classifiers for datasets with missing values,'' \emph{Communications Medicine}, vol.~3, no.~1, 2023.

\bibitem{Kim23}
S.~Kim, H.~Kim, E.~Yun, H.~Lee, J.~Lee, and J.~Lee, ``Probabilistic imputation for time-series classification with missing data,'' in \emph{Proceedings of the 40th International Conference on Machine Learning}, 2023.

\bibitem{Murphy12}
K.~P. Murphy, \emph{Machine Learning: A Probabilistic Perspective}.\hskip 1em plus 0.5em minus 0.4em\relax The MIT Press, 2012.

\bibitem{elkan2001costsensitive}
C.~Elkan, ``The foundations of cost-sensitive learning,'' in \emph{International Joint Conference on Artificial Intelligence}, vol.~17, 2001.

\bibitem{jebara12}
T.~Jebara, \emph{Machine Learning: Discriminative and Generative}.\hskip 1em plus 0.5em minus 0.4em\relax Springer New York, NY, 2012.

\bibitem{friedman97}
N.~Friedman, D.~Geiger, and M.~Goldszmidt, ``Bayesian network classifiers,'' \emph{Machine Learning}, vol.~29, no. 2--3, pp. 131--163, 1997.

\bibitem{marfoq23}
O.~Marfoq, G.~Neglia, L.~Kameni, and R.~Vidal, ``Federated learning for data streams,'' in \emph{Proceedings of The 26th International Conference on Artificial Intelligence and Statistics}, 2023.

\bibitem{domingos1997optimalityNB}
P.~Domingos and M.~Pazzani, ``On the optimality of the simple {B}ayesian classifier under zero-one loss,'' \emph{Machine learning}, vol.~29, pp. 103--130, 1997.

\bibitem{rish2001empiricalNB}
I.~Rish \emph{et~al.}, ``An empirical study of the naive {B}ayes classifier,'' in \emph{IJCAI workshop on empirical methods in artificial intelligence}, vol.~3, 2001.

\bibitem{zhang2004optimalityNB}
H.~Zhang, ``The optimality of naive {B}ayes,'' in \emph{Proceedings of the Seventeenth International Florida Artificial Intelligence Research Society Conference}, 2004.

\bibitem{greiner02}
R.~Greiner and W.~Zhou, ``Structural extension to logistic regression: Discriminative parameter learning of belief net classifiers.'' 2002.

\end{thebibliography}

\end{document}